\documentclass[11pt,a4paper]{article}

\usepackage{polyu-vclab}

\usepackage{lipsum}                
\usepackage{amsmath}
\usepackage{booktabs}
\usepackage{multirow}
\usepackage[numbers,sort&compress]{natbib}
\usepackage{nicematrix}
\usepackage{hyperref}
\usepackage{cleveref}
\usepackage{subcaption}

\setEyebrow{PolyU VCLab\,\textbullet\,Preprint 2026}

\setReportTag{Visual Computing Lab\,\textperiodcentered\,The Hong Kong Polytechnic University}

\papertitle{DepthMaster: Unified Monocular Depth Estimation for Perspective and Panoramic Images}

\paperauthors{%
  Pengfei Wang\quad
  Shihao Wang\quad
  Liyi Chen\quad
  Zhiyuan Ma\quad
  Guowen Zhang\quad
  Lei Zhang\correspondmark%
}

\paperaffil{%
  The Hong Kong Polytechnic University
}

\papernotes{%
\href{mailto:pengfei.wang@connect.polyu.hk }{pengfei.wang@connect.polyu.hk } \quad
  \correspondmark\,Corresponding author.%
}

\paperbadges{%
  \vclabbadgesolid{Project Page}{https://polyu-vclab.github.io/DepthMaster-page}\;%
  \vclabbadgesolid{Code}{https://github.com/PolyU-VCLab/DepthMaster}\;%
  \vclabbadgesolid{Model}{https://huggingface.co/VCLab-PolyU/DepthMaster}%
}

\begin{document}

\maketitleVCLab

\begin{vclabAbstract}
\noindent\textbf{Abstract.}\;
While monocular depth estimation has achieved significant progress, achieving generalized metric depth estimation for both narrow field-of-view (FoV) perspectives and $360^\circ$ panoramas remains an unsolved challenge.
Existing methods are often tailored to specific camera types and struggle to produce accurate metric depth that generalizes across diverse settings.
This limitation stems from two key challenges: the inherent geometric discrepancy between perspective and panoramic cameras, and the scarcity of panoramic training data with metric annotations.
In this work, we introduce \textbf{DepthMaster}, a unified metric depth estimation framework.
Rather than employing specialized networks to learn spherical distortions, we reformulate the problem by decomposing panoramic images into overlapping perspective patches. 
Crucially, distinct from prior projection-based methods that rely on ad-hoc architectural modifications to handle boundaries, we introduce a novel Correspondence Consistency Loss (CCL) and inject virtual projection cameras as geometric priors, allowing us to seamlessly stitch the patches while avoiding specialized operators and keeping the backbone largely compatible with standard Transformer designs.
This strategy also resolves the geometric differences by unifying all inputs into a canonical perspective representation, and effectively circumvents data scarcity by directly unlocking powerful metric priors from vast perspective datasets.
Trained on a mixed dataset that contains only one panorama dataset, DepthMaster achieves state-of-the-art zero-shot performance on 13 diverse datasets, outperforming \textbf{not only universal methods but also leading specialist models in both perspective and panoramic domains}.
\end{vclabAbstract}


\section{Introduction}
\label{sec:intro}
\vspace{-2mm}

Monocular depth estimation (MDE), particularly metric depth estimation, is a fundamental task in computer vision, which has wide applications in 3D scene understanding, autonomous navigation, augmented reality, and robotics~\cite{Huang2DGS2024,long2024wonder3d,wang2024shape,lei2024mosca,SpatialTracker,yurtsever2020survey,hu2023planning,wang2024open,chen2026fast,wang2026one2scene,chen2026omni}. In recent years, MDE has achieved remarkable progress, largely driven by deep learning models trained on large-scale, diverse, and high-quality data~\cite{yin2023metric3d, hu2024metric3d, yang2024depthanything, yang2024depth2, ranftl2020midas,wang2025moge}. State-of-the-art (SOTA) methods \cite{wei2021leres,piccinelli2024unidepth,yang2024depthanything,yang2024depth2,Bochkovskii2024,wang2025moge2} have demonstrated impressive zero-shot generalization performance, capable of recovering high-fidelity metric depth from diverse in-the-wild perspective images.

\begin{figure*}[t!]
    \centering
\includegraphics[width=\linewidth]{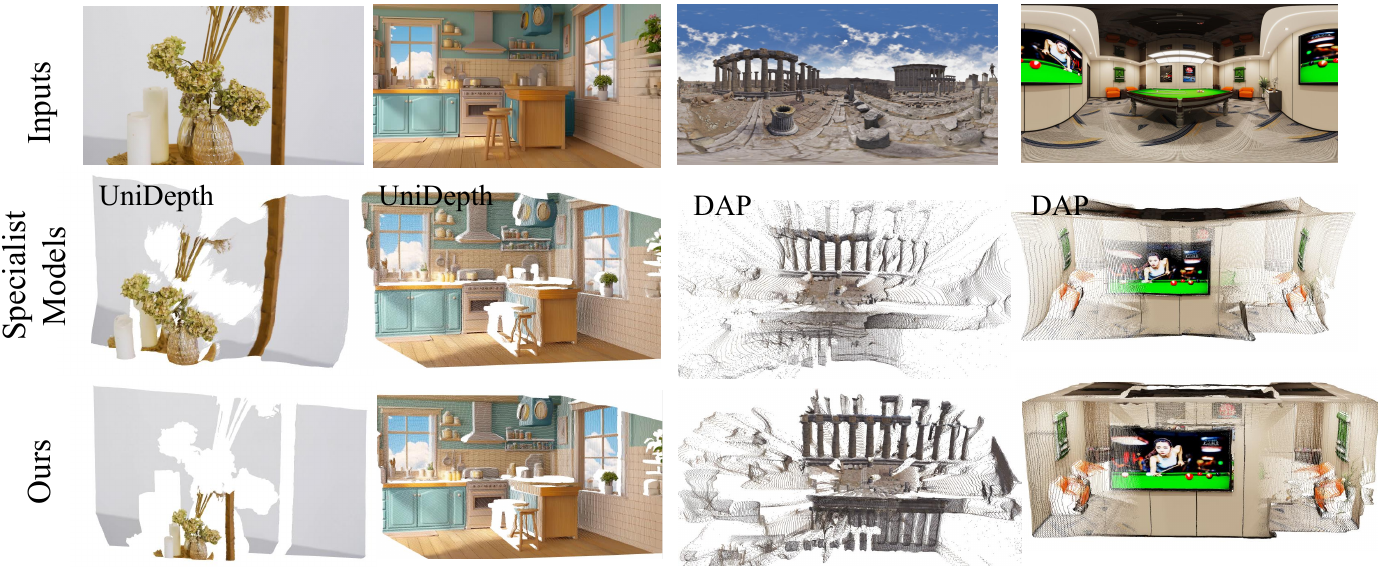} 
\vspace{-5mm}
    \caption{\textbf{DepthMaster enables high-fidelity zero-shot metric depth estimation.} We visualize the 3D point clouds reconstructed from diverse inputs. While specialist models—such as UniDepth~\cite{piccinelli2024unidepth} for perspective images and Depth Any Panoramas (DAP)~\cite{lin2025dap} for panoramas—often struggle with global geometric structures or local fine details, our unified DepthMaster produces superior 3D structures and sharper details across both camera types. We provide more interactive visualizations on our project page: \url{https://polyu-vclab.github.io/DepthMaster-page}.}
    \label{fig:teaser}
    \vspace{-4mm}
\end{figure*}

However, the success of these SOTA models is predominantly confined to standard pinhole cameras, and their performance degrades dramatically on $360^\circ$ panoramic images due to severe equirectangular projection (ERP) distortions. While specialist models have been developed for specific domains (e.g., Depth Any Panoramas (DAP)~\cite{lin2025dap} for panoramas), they still exhibit noticeable limitations in real-world scenarios. As illustrated by the 3D point cloud reconstructions in \Cref{fig:teaser}, these domain-specific specialists often struggle to preserve accurate global geometric structures and fine details. For panoramic specialists, this vulnerability stems primarily from the acute scarcity of panoramic training data, which severely restricts their generalization capabilities. More critically, their specialized, distortion-aware architectures are inherently incompatible with standard perspective inputs, precluding a unified solution to perspective and panorama images.

Consequently, developing a universal metric MDE model faces two intertwined challenges. (1) The first challenge is geometric generalization~\cite{piccinelli2025unik3d,guo2025depth}. A universal model must reconcile the linear projection of pinhole cameras with the non-linear, spherical projection of panoramic cameras without adopting distinct processing pipelines. (2) Another, and perhaps more fundamental, obstacle is data imbalance. While modern perspective MDE methods have flourished by leveraging massive, high-quality datasets, comparable resources for panoramic images are acutely lacking~\cite{wang2024depth,Wang2020BiFuseM3}. The key to overcoming this scarcity lies in effectively transferring the rich perspective priors to the panoramic domain. However, existing panoramic methods fail to achieve this because their ad-hoc architectural modifications prevent them from directly inheriting weights and priors from foundation models trained on standard perspective data.

To address these challenges, we introduce \textbf{DepthMaster}, a unified monocular metric depth estimation framework. While canonicalizing panoramas into local perspective patches is a common practice to mitigate spherical distortions, prior methods often rely on ad-hoc architectural modifications (e.g., cube convolution~\cite{lichy2024fova}) to fuse these patches and handle seam artifacts. Such specialized designs inherently disrupt the standard network structure, hindering the direct transfer of powerful perspective priors learned from large-scale data. Our core innovation lies in  unifying panoramic and perspective depth estimation without altering the standard Transformer architecture. Specifically, we project the panoramic image into tangent patches with a $2.5^\circ$ FoV overlap and inject the spatial layouts of the virtual projection cameras as geometric priors. To ensure global consistency and seamless fusion across patch boundaries, we propose a novel Correspondence Consistency Loss (CCL) as a soft constraint. This loss-based design elegantly substitutes architectural modifications, perfectly preserving the pure backbone. By unifying inputs into this canonical representation and leveraging virtual camera conditioning, DepthMaster accommodates both standard perspective images and panoramic patches, maximizing the utilization of pre-trained perspective priors.

To validate our approach, we perform comprehensive zero-shot metric evaluations across multiple datasets, encompassing both perspective and $360^\circ$ panoramic domains. The results demonstrate that DepthMaster not only decisively outperforms existing universal methods, such as Unik3D~\cite{piccinelli2025unik3d} and DAC~\cite{guo2025depth}, but also excels specialist models in their dedicated domains. Specifically, it outperforms leading metric estimators like Depth Anything~\cite{yang2024depthanything, yang2024depth2} on perspective imagery and HUSH~\cite{lee2025hush} on panoramic scenes. This dual superiority establishes DepthMaster as the new SOTA in unified monocular metric depth estimation.
Our contributions are summarized as follows:

\begin{itemize} 
    \item We propose \textbf{DepthMaster}, a unified monocular metric depth estimation framework. By injecting the spatial layouts of virtual projection cameras as geometric priors, we effectively bridge the geometric gap between perspective and panoramic domains without disrupting the standard Transformer architecture.

    \item We introduce a Correspondence Consistency Loss (CCL) as a soft constraint applied to overlapping patches (e.g., $2.5^\circ$ FoV overlap). This loss elegantly enforces global consistency and local patch fusion, avoiding ad-hoc architectural modifications and maximizing the inheritance of powerful pre-trained perspective priors.

    \item DepthMaster achieves \textbf{SOTA zero-shot performance} on 13 diverse datasets, outperforming both universal methods and specialist models on their respective benchmarks.
\end{itemize}

\section{Related Work}
\label{sec:relwork}
\vspace{-1mm}

\noindent{}\textbf{Monocular Depth Estimation.}
Early CNN-based methods \cite{eigen2014depth,fu2018deep,yuan2022neural} focus on in-domain data and lack generalization performance. To enhance zero-shot generalization, recent works can be partitioned into two main categories. 
The first scales up model capacity and trains on massive, diverse datasets~\cite{ranftl2021vision,yin2021virtual,ranftl2020midas}, such as the recent SOTA estimators DepthAnything~\cite{yang2024depth1, yang2024depth2}, Metric3D~\cite{yin2023metric3d, hu2024metric3d}, and MoGe~\cite{wang2025moge, wang2025moge2} trained with millions of samples. However, these methods are designed basically for perspective camera images. 
The second category re-purposes pre-trained diffusion priors~\cite{xu2024diffusion,ke2024repurposing,fu2024geowizard} for depth estimation.  
Our work aligns with the first category, but expands its scope to encompass diverse camera types. By canonicalizing panoramas into overlapping perspective patches and fusing them via a pure loss-based constraint (CCL), we elegantly extend standard perspective foundation models to the panoramic domain without architectural disruption.

\noindent{}\textbf{Depth Estimation for Panorama.}
How to handle ERP distortions is the primary challenge for panoramic depth estimation~\cite{sun2021hohonet,ai2024elite360d,yan2025spherefusion,lee2025hush}. While early methods adapted convolutional kernels~\cite{deformcnn360,xiong2024efficient} or applied Transformers globally~\cite{omnifusion,panoformer,li2025depth}, recent approaches project ERP into less distorted representations. 
However, methods like BiFuse~\cite{Wang2020BiFuseM3} (dual-branch ERP-cubemap fusion), FoVA-Depth~\cite{lichy2024fova} (customized CubeConv), and OmniFusion~\cite{omnifusion} (18 tangent images) rely on specialized, ad-hoc architectural modifications. Such designs inherently disrupt the standard network backbone, hindering joint training with large-scale perspective data and blocking the direct inheritance of powerful pre-trained perspective priors.
In contrast, our approach resolves boundary and fusion issues using virtual camera conditioning together with a soft constraint (CCL) without structural modifications, effectively exploiting perspective priors and boosting panoramic performance.

\noindent{}\textbf{Unified Monocular Depth Estimation.}
Recent advances seek to reconcile diverse imaging geometries. UniK3D~\cite{piccinelli2025unik3d} constructs a spherical 3D representation, but its reliance on scarce panoramic data limits generalization. Conversely, DAC~\cite{guo2025depth} establishes ERP as the canonical representation by projecting perspective images into ERP patches. This strategy suffers from two drawbacks: it requires ground-truth camera intrinsics during inference, precluding in-the-wild application, and forcing perspective data into the distorted ERP domain impedes the transfer of perspective priors.
In contrast, we unify all inputs within the canonical perspective domain. We decompose panoramas into overlapping perspective patches and inject known virtual spatial layouts, requiring zero ground-truth camera metadata. By preserving the native standard backbone and enforcing global consistency via CCL, our approach overcomes data scarcity and achieves SOTA performance across both benchmarks.

\section{Method}
\label{sec:method}
\vspace{-2mm}

We introduce \textbf{DepthMaster}, a unified metric depth estimation framework for perspective and 
$360^\circ$ cameras. Instead of relying on task-specific operators tailored to cubemap topology, we resolve geometric discrepancies by first canonicalizing any input image into a set of perspective-like patches (\cref{sec:canonicalization}), then injecting their virtual spatial layouts as geometric conditions and processing them with a standard Transformer network (\cref{ssec:method:framework}). Finally, we fuse these patches using our proposed CCL loss (\cref{ssec:method:ccl}), which acts as a soft geometric constraint to preserve backbone purity.
  
\subsection{Canonicalization to Perspective Patches} \label{sec:canonicalization}

The primary challenge in unifying depth estimation for perspective and $360^\circ$ images lies in their distinct geometric projections—linear versus spherical. To bridge this gap, we employ a geometric canonicalization strategy that transforms spherical signals into standard perspective representations, enabling the direct transfer of powerful metric priors within a unified framework.

\begin{figure*}[!tb]
    \centering
    \includegraphics[width=0.99\textwidth]{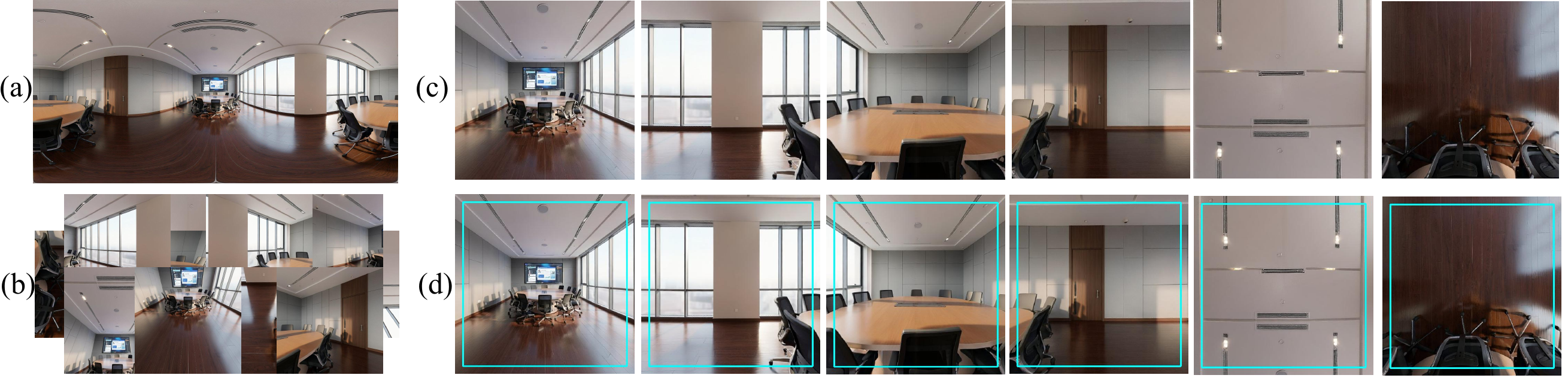}
    \vspace{-2mm}
    \caption{\textbf{Comparison of different projection strategies.} (a) Equirectangular. (b) Tangent projection with 18 overlapping patches~\cite{omnifusion}. (c) Standard 
$90^\circ$ 
  Cubemap~\cite{Wang2020BiFuseM3}. (d) Our proposed  
$95^\circ$ Cubemap with \(2.5^\circ\) overlap on each side.
    }
   \label{fig:cube}
   \vspace{-4mm}
\end{figure*}

Rather than learning spherical distortions directly or employing complex multi-tangent schemes (e.g., 18 images~\cite{omnifusion}), we adopt a simple, standard multi-view formulation. Using tangent plane projection, we unfold the panoramic image into six canonical perspective patches. 
As illustrated in \Cref{fig:cube}, standard $90^\circ$ cubemaps suffer from severe boundary discontinuities. To address this, prior methods like BiFuse~\cite{Wang2020BiFuseM3} and FoVA-Depth~\cite{lichy2024fova} are forced to introduce customized architectural modifications (e.g., dual-branch fusion or specialized CubeConv operators) to explicitly handle the topology. However, these specific designs largely alter the network structure, preventing direct inference on perspective images and blocking the reuse of pre-trained weights.

To avoid disrupting the backbone while addressing boundary artifacts, we consider expanding the FoV of each patch. However, excessive FoV expansion introduces distortion, whereas expanding by only $2.5^\circ$ introduces almost none, as illustrated in \Cref{fig:cube}.
The virtual focal length is accordingly adjusted to $f = (w/2) / \tan(95^\circ/2)$ for a patch width $w$, establishing a shared intrinsic matrix $K$.
This minimal FoV expansion generates the critical overlapping regions required for our CCL loss (\cref{ssec:method:ccl}). By standardizing the inputs into perspective patches with overlaps, this strategy bypasses domain-specific operators, better preserves compatibility with pre-trained perspective priors, and enables a unified inference pipeline across both camera types.

\subsection{Framework Architecture} \label{ssec:method:framework}
\begin{figure*}[!tb]
    \centering
    \includegraphics[width=0.98\textwidth]{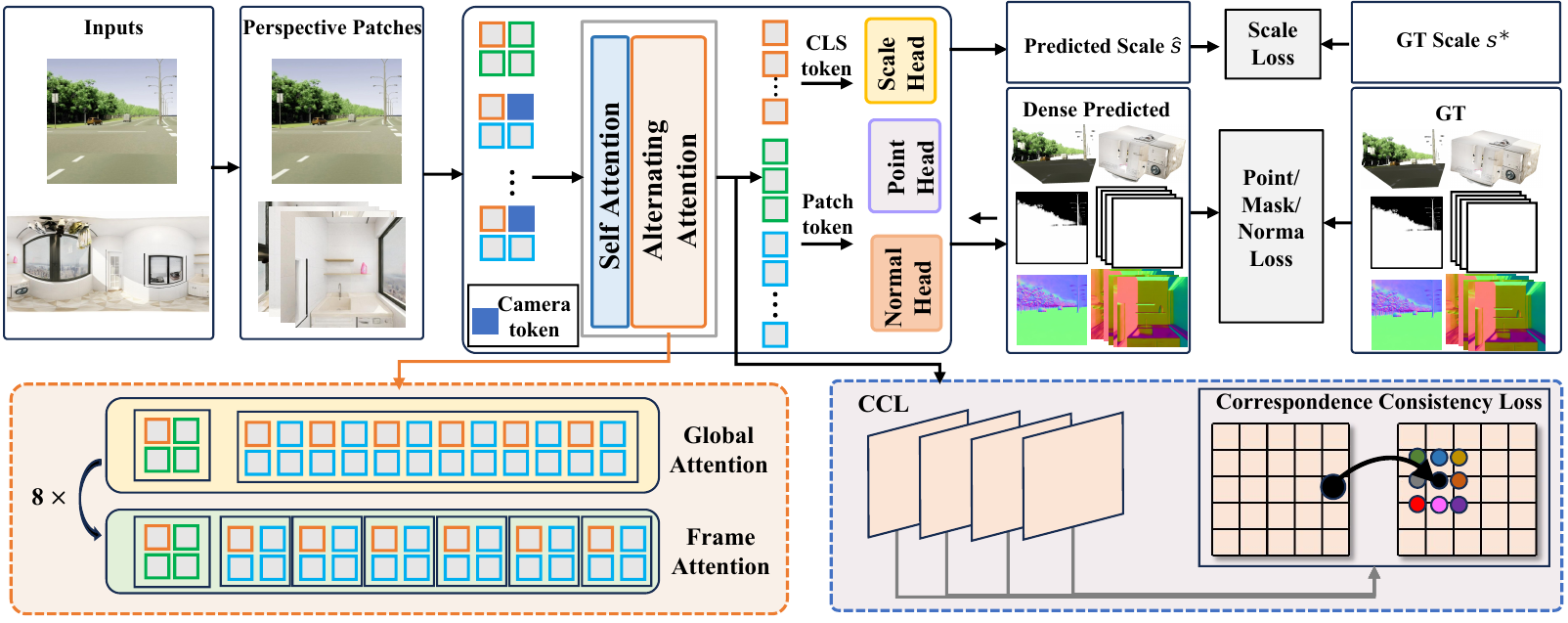}
    \caption{\textbf{The pipeline of DepthMaster.} Perspective and panoramic images are unified as perspective patches and processed by a shared transformer with virtual camera conditioning, while CCL explicitly enforces consistency across overlapping patches during training. The CLS token for perspective inputs, or pooled CLS tokens for panoramic inputs, is used to predict a shared global scale.}
   \label{fig:pipeline}
   \vspace{-4mm}
\end{figure*}

The architecture of DepthMaster is illustrated in \Cref{fig:pipeline}, which mainly consists of a transformer backbone with camera conditioning, a dense prediction head, and a scale prediction head. 

\noindent\textbf{Transformer Backbone.}
Our architecture is built upon DINOv2~\cite{oquab2023dinov2} ViT-L/14, a widely adopted backbone for depth estimation tasks that comprises 24 Transformer blocks. 
Following a design of DA3 \cite{lin2025depth}, the initial \(8\) layers retain the standard DINOv2 self-attention mechanism. 
Starting from the \(9\)-th layer, we introduce alternating attention into the remaining \(16\) blocks. The alternating attention is achieved entirely by reshaping the input tensor dimensions. 
The frame-wise attention captures intra-patch structure, whereas global attention establishes interactions across patches, thereby facilitating spatial correspondence and geometric coherence. 
Our canonicalization process generates $N$ perspective patches: a single patch ($N{=}1$) for standard perspective images or six patches ($N{=}6$) for panoramic inputs. 
In the single-patch scenario, global attention naturally reduces to standard self-attention, ensuring unified training for both inputs.

\noindent\textbf{Virtual Camera Conditioning.}
For panoramic inputs, we leverage the virtual camera extrinsics and intrinsics used in cubemap projection (\cref{sec:canonicalization}) as conditioning signals to provide geometric priors.

The world-to-camera extrinsic matrix $\mathbf{T} \in SE(3)$ is first converted to a camera-to-world representation and parameterized as a 7-dimensional vector $\mathbf{p} = [\mathbf{t}; \mathbf{q}] \in \mathbb{R}^7$, where $\mathbf{t} \in \mathbb{R}^3$ is the translation and $\mathbf{q} \in \mathbb{R}^4$ is the unit quaternion encoding rotation. This vector is embedded via a two-layer MLP with SiLU activation:
\begin{equation}
    \mathbf{e}_{\text{pose}} = \text{MLP}_{\text{pose}}(\mathbf{p}) \in \mathbb{R}^d,
\end{equation}
and prepended to the token sequence as a special conditioning token.

The camera intrinsics are summarized as a 4-dimensional vector $\mathbf{r} = [f_x/W, f_y/H, c_x/W, c_y/H]$, encoding the normalized focal lengths and principal point, which is embedded via a two-layer MLP:
\begin{equation}
    \mathbf{e}_{\text{ray}} = \text{MLP}_{\text{ray}}(\mathbf{r}) \in \mathbb{R}^d,
\end{equation}
and prepended as another special token. For perspective image inputs, no camera conditioning is applied, allowing the model to jointly train on perspective images without requiring camera metadata.

\noindent{}\textbf{Dense Prediction Heads.} 
We have two dense prediction heads sharing the same Dense Prediction Transformer (DPT) architecture~\cite{ranftl2021vision}, which takes multi-scale features from layers $\{11, 15, 19, 23\}$ of the backbone. 
The point map head outputs a 4-channel dense prediction \(\hat{\mathbf{P}} \in \mathbb{R}^{H \times W \times 4}\). The first three channels represent the 3D point map in world coordinates for panoramic inputs or affine camera space for perspective inputs. The fourth channel predicts a single-channel validity mask \(\hat{\mathbf{M}} \in [0,1]^{H \times W}\) indicating non-sky regions. For the 3D point map, we supervise it using affine-invariant point cloud loss. For each sphere scale \(\alpha\), we sample anchor points \(\mathcal H_\alpha\) and compute:
\begin{equation}
    \mathcal L_{S(\alpha)} = \sum_{j\in \mathcal H_\alpha} \sum_{i\in { S}_j} \frac{1}{z_i}\left\| s_j^* \hat{\mathbf p}_i+\mathbf t_j^*- {\mathbf p}_i\right\|_1,
\end{equation}
where we apply the ROE alignment solver \cite{wang2025moge} to obtain \((s_j^*,\mathbf t_j^*)\). We use three scales: \(\alpha \in \{1/4, 1/16, 1/64\}\), denoted as \(\mathcal L_{S_1}\), \(\mathcal L_{S_2}\), and \(\mathcal L_{S_3}\). For the mask channel, we supervise it with the mask BCE loss:
\begin{equation}
    \mathcal L_M = \|\hat{\mathbf M} - (1-{\mathbf M}_\text{inf})\|_2^2,
\end{equation}
where \({\mathbf M}_\text{inf}\) is the infinity mask label obtained from ground truth for synthetic data or SegFormer~\cite{xie2021segformer} for outdoor scenes.
The surface normal head outputs an \(\ell_2\)-normalized 3-channel normal map \(\hat{\mathbf{N}} \in \mathbb{R}^{H \times W \times 3}\) with unit vectors \(\hat{\mathbf{n}} = \frac{\mathbf{x}}{\|\mathbf{x}\|_2}\). We supervise it with the following normal loss:
\begin{equation}
    \mathcal L_N = \sum_{i\in {\mathcal M}}\angle (\hat{\mathbf n}_i, \mathbf n_i),
\end{equation}
where \(\hat{\mathbf n}_i\) is the predicted unit normal from the normal head, and \(\angle (\cdot,\cdot)\) measures angular difference.

\noindent{}\textbf{Scale Prediction Head.}
In metric depth prediction approaches such as Metric3D~\cite{hu2024metric3d} and MoGe v2~\cite{wang2025moge2}, scale is typically estimated separately. To avoid compromising the accuracy of relative geometry, we introduce a dedicated branch with independent supervision for scale prediction:
\begin{equation}
\mathcal{L}_{s} = \left\| \log(\hat{s}) - \text{stopgrad}(\log(s^*)) \right\|_2^2,
\end{equation}
where $\log(\hat{s})$ denotes the predicted metric scale in log space, and $s^*$ is the optimal scale computed on-the-fly between the predicted scale-invariant depth map $\hat{d}$ and the ground truth. To ensure strict scale consistency across multi-patch panoramic inputs, we perform average pooling over the CLS tokens of all patches to form a unified global representation before passing it to the scale predictor. By predicting a single, globally shared scale factor and multiplying it with the scale-invariant point maps, our framework inherently avoids scale discontinuities or drift at the patch boundaries.

\subsection{Correspondence Consistency Loss}
\label{ssec:method:ccl}

Although global attention facilitates information exchange across patches, it does not explicitly enforce consistency within overlapping regions. To address this limitation, we introduce a novel loss function, namely Correspondence Consistency Loss (CCL), grounded in dense patch correspondences.

\noindent{}\textbf{Building Correspondences.}
For each pair of overlapping patches $(i, j)$, we construct a dense correspondence field based on their relative orientations.  
Each patch is treated as a virtual perspective camera sharing the same intrinsic matrix $K$ but having its own rotation matrix $R_k$,  
which maps points from the camera coordinate system to the world coordinate system.  
Since we only consider directions at infinity, the mapping between patches $i$ and $j$ can be determined by:
\begin{equation}
\label{eq:homography}
\mathbf{H}_{ij} = K\, R_j^{-1}\, R_i\, K^{-1},
\end{equation}
which warps a pixel $\mathbf{p}_i$ in patch $i$ to its corresponding location
$\mathbf{p}_j \sim \mathbf{H}_{ij}\mathbf{p}_i$ in patch $j$.  
Because $\mathbf{p}_j$ often lies at non-integer coordinates, interpolating directly on the final predicted point clouds would introduce significant errors. Instead, applying CCL at the feature level enforces rich, high-dimensional structural alignment, making it a more robust choice. Thus, the feature representation $\mathbf{F}_j^m(\mathbf{p}_j)$ at backbone layer $m$ is obtained via differentiable bilinear interpolation from the feature map $\mathbf{F}_j^m$.

\noindent{}\textbf{Loss Formulation.} 
The CCL is applied to multi-scale features from layers $\{11, 15, 19, 23\}$ of the backbone. It aggregates the $L_2$ distance between the corresponding feature vectors over all valid correspondences, overlapping pairs $\mathcal{O}$, and selected backbone layers:
\begin{equation}
\mathcal{L}_{\text{CCL}} =
\sum_{m=1}^{M}
\sum_{(i,j)\in\mathcal{O}}
\frac{1}{|\Omega_{ij}|}
\sum_{(\mathbf{p}_i,\mathbf{p}_j)\in\Omega_{ij}}
\bigl\|
\mathbf{F}_i^m(\mathbf{p}_i) -
\mathbf{F}_j^m(\mathbf{p}_j)
\bigr\|_2^2,
\end{equation}
where $M$ is the number of backbone layers to which the loss is applied. 
By enforcing such correspondence constraints during learning, we are able to learn geometry-aware representations that are geometrically consistent in the overlapping boundary regions.

\section{Experiments}
\label{sec:exp}
\subsection{Experimental Settings}
\label{exp: detail}

 \noindent\textbf{Experiments Setup}. Our model is trained on 15 datasets, including 14 perspective datasets and 1 panoramic (i.e., Structured3D \cite{Structured3D}). For real-world data, due to inherent noise (e.g., incompleteness, blurred object boundaries, and severe flying pixels), we refine the ground-truth depth maps. To evaluate the zero-shot generalization capabilities of our DepthMaster model, we test it on 13 unseen datasets (10 perspective and 3 panoramic). It is worth mentioning that \textbf{our model is trained on a significantly smaller data scale compared to existing models} such as MoGe~\cite{wang2025moge}, DepthAnything~\cite{yang2024depthanything}, and the unified model UniK3D~\cite{piccinelli2025unik3d} (see Appendix~\ref{Appendix:data_scale}). {Detailed dataset statistics, implementation details, and real data refinement strategies are provided in Appendix~\ref{Appendix:data} and \ref{Appendix:Details}, respectively.

\noindent\textbf{Evaluation Details}. We conduct comprehensive evaluations of DepthMaster's zero-shot generalization capabilities across both perspective and panoramic images using a \textbf{single pre-trained model without any domain-specific fine-tuning}. For perspective images, we follow the MoGe protocol~\cite{wang2025moge}. For panoramic images, evaluations are performed on the full ERP maps following the DA\(^{2}\)~\cite{li20252} protocol. While our method directly outputs a unified 3D point cloud in the world coordinate system, to ensure a fair comparison with existing baselines, we reproject the predicted \(95^\circ\) FoV cubemap back into a 2D ERP depth map format, applying a simple weighted average to merge the overlapping regions. Comprehensive evaluation protocols and metrics are detailed in Appendix~\ref{Appendix:eval}.

\subsection{Main Results}

\vspace{+1mm}
\noindent\textbf{Results on Panoramic Images.}
We evaluate DepthMaster on three panoramic benchmarks: Stanford2D3D~\cite{stanford2d3ds}, Matterport3D~\cite{chang2017matterport3d}, and PanoSUNCG~\cite{PanoSUNCG}. As shown in \Cref{tab:main}, DepthMaster achieves remarkable zero-shot performance across all three evaluation protocols, surpassing both specialized and unified models.
For scale-invariant depth, DepthMaster substantially outperforms the recent model DA$^{2}$~\cite{li2025depth}, reducing the average AbsRel from 6.05 to 4.77 and boosting $\delta_1$ accuracy from 95.63\% to 96.19\%.
For affine-invariant depth, DepthMaster further outperforms the SOTA specialized model DA360~\cite{jiang2025depth}, cutting the average AbsRel from 6.39 to 4.58 and improving $\delta_1$ from 94.53\% to 96.33\%. Critically, DepthMaster's zero-shot performance on Matterport3D surpasses UniK3D (5.79 vs. 8.14 AbsRel) by a large margin, although UniK3D is explicitly trained on this dataset.
For metric depth, DepthMaster also consistently outperforms the SOTA metric depth model DAP~\cite{lin2025dap} (7.25 vs. 11.79 AbsRel).
It is worth mentioning that DepthMaster achieves these results by \textbf{training only on a single panoramic dataset}, demonstrating its exceptional data efficiency.

\noindent\textbf{Results on Perspective Images.}
We further evaluate DepthMaster's performance on ten diverse perspective datasets: NYUv2~\cite{Silberman2012nyuv2}, KITTI~\cite{Uhrig2017kitti}, ETH3D~\cite{Schops2019ETH3D}, iBims-1~\cite{ibim1_1,ibims1_2}, GSO~\cite{downs2022googlescannedobjects}, Sintel~\cite{Butler2012sintel}, DDAD~\cite{ddadpacking}, DIODE~\cite{diode_dataset}, Spring~\cite{Mehl2023Spring}, and HAMMER~\cite{jung2023hammer}. We benchmark our method against strong specialist baselines across both relative and metric geometry settings. As shown in Table~\ref{tab:quantitative_comparison_relative_and_metric}, \textbf{DepthMaster achieves the best overall average rank (1.31), outperforming the previous strongest method, MoGe 2 (2.22), by a large margin}. Specifically, our method delivers the best average results in scale-invariant point prediction, affine-invariant point prediction, metric depth prediction, scale-invariant depth prediction, affine-invariant depth prediction, and affine-invariant disparity prediction. For metric point prediction, it achieves the top $\delta_1$ score (93.9\%), while remaining competitive in AbsRel (8.27). These results demonstrate DepthMaster's strong generalization across diverse perspective scenes, despite using only one-fourth the training data of recent unified models (see Appendix~\ref{Appendix:data_scale}).

\setlength{\tabcolsep}{2pt}
\begin{table*}[t!]
\centering
\fontsize{8.0pt}{9.0pt}\selectfont
\caption{
Quantitative results on three panorama datasets. 
The \textcolor{red}{best} and \textcolor{blue}{second best} results are highlighted in \textcolor{red}{red} and \textcolor{blue}{blue}, respectively. 
\textcolor{lightgray}{Gray numbers} denote models trained on this dataset. Unavailable/unsupported metrics are marked with `\textcolor{lightgray}{-}'.
Methods marked with $^{\dagger}$ are in-domain models trained on the corresponding panorama datasets, and are listed for reference only (not ranked).
Results in the Scale-invariant depth map section are taken from {DA}$^{2}$~\cite{li2025depth}.
Our model surpasses not only other unified models but also all specialized models designed specifically for panoramic imagery.
}
\vspace{-2mm}
\resizebox{\textwidth}{!}{
\begin{NiceTabular}{c|cccc|cccc|cccc|cc}
\toprule
\multirow{2}{3em}{Method} &
\multicolumn{4}{c}{Stanford2D3D} &
\multicolumn{4}{c}{Matterport3D} &
\multicolumn{4}{c}{PanoSUNCG} &
\multicolumn{2}{c}{\textit{Avg.}}\\

\cmidrule(lr){2-5}
\cmidrule(lr){6-9}
\cmidrule(lr){10-13}
\cmidrule(lr){14-15}
&
AbsRel$\downarrow$ & RMSE$\downarrow$ & $\delta_1\uparrow$ & $\delta_2\uparrow$ &
AbsRel$\downarrow$ & RMSE$\downarrow$ & $\delta_1\uparrow$ & $\delta_2\uparrow$ &
AbsRel$\downarrow$ & RMSE$\downarrow$ & $\delta_1\uparrow$ & $\delta_2\uparrow$ &
AbsRel$\downarrow$ & $\delta_1\uparrow$ \\

\midrule
\multicolumn{15}{c}{\rule{0pt}{2.4ex}Scale-invariant depth map\rule[-1.0ex]{0pt}{0pt}}\\[2pt]
\hline
OmniDepth$^{\dagger}$~\cite{zioulis2018omnidepth} & 19.96 & 61.52 & 68.77 & 88.91 & 29.01 & 76.43 & 68.30 & 87.94 & 11.43 & 37.10 & 87.05 & 93.65 & 20.13 & 74.71 \\
BiFuse$^{\dagger}$~\cite{wang2020bifuse} & 12.09 & 41.42 & 86.60 & 95.80 & 20.48 & 62.59 & 84.52 & 93.19 & 5.92 & 25.96 & 95.90 & 98.23 & 12.83 & 89.01 \\
BiFuse++$^{\dagger}$~\cite{wang2022bifuse++} & 11.17 & 37.20 & 87.83 & 96.49 & 14.24 & 51.90 & 87.90 & 95.17 & 5.24 & 24.77 & 96.30 & 98.35 & 10.22 & 90.68 \\
UniFuse$^{\dagger}$~\cite{jiang2021unifuse} & 11.14 & 36.91 & 87.11 & 96.64 & 10.63 & 49.41 & 88.97 & 96.23 & 5.28 & 27.04 & 95.91 & 98.25 & 9.02 & 90.66 \\
PanoFormer$^{\dagger}$~\cite{shen2022panoformer} & 11.31 & 35.57 & 88.08 & 96.23 & 9.04 & 44.70 & 88.16 & 96.61 & 5.34 & 18.90 & 94.87 & 98.83 & 8.56 & 90.37 \\
HRDFuse$^{\dagger}$~\cite{ai2023hrdfuse} & 9.35 & 31.06 & 91.40 & 97.98 & 9.67 & 44.33 & 91.62 & 96.69 & 6.90 & 27.44 & 92.15 & 97.42 & 8.64 & 91.72 \\
DepthAnywhere$^{\dagger}$~\cite{wang2024depth} & 11.80 & 35.10 & 91.00 & 97.10 & 8.50 & -- & 91.70 & 97.60 & -- & -- & -- & -- & -- & -- \\
PanDA~\cite{cao2025panda} & 48.44 & 53.06 & 33.92 & 51.33 & 37.10& 101.5& 42.51& 67.29& 34.73& 79.69& 44.49& 71.45& 40.09& 40.31\\

 UniK3D~\cite{piccinelli2025unik3d} &  11.31&   {19.72}&  88.94& 95.33&
\textcolor{lightgray}{9.66} &   \textcolor{lightgray}{32.66}&   \textcolor{lightgray}{93.00}&   \textcolor{lightgray}{98.58}& 
11.46& 25.38& 90.18& 98.02& 11.39 & 89.56 \\
{DA}$^{2}$~\cite{li2025depth} &
  \textcolor{blue}{6.09} & \textcolor{blue}{19.09} & \textcolor{blue}{95.94} & \textcolor{blue}{98.51}&
 \textcolor{blue}{6.83} & \textcolor{blue}{27.51} & \textcolor{blue}{94.89} & \textcolor{blue}{97.71}&
 \textcolor{blue}{5.22} & \textcolor{blue}{19.31} & \textcolor{blue}{96.06} & \textcolor{blue}{98.28} & \textcolor{blue}{6.05} & \textcolor{blue}{95.63} \\ 
{DepthMaster (Ours)} &
\textcolor{red}{4.81} & \textcolor{red}{17.63} & \textcolor{red}{96.15} & \textcolor{red}{98.56} &
\textcolor{red}{5.95} & \textcolor{red}{24.71} & \textcolor{red}{94.92} & \textcolor{red}{98.13} &
\textcolor{red}{3.54} & \textcolor{red}{15.46} & \textcolor{red}{97.50} & \textcolor{red}{98.81} &
\textcolor{red}{4.77} & \textcolor{red}{96.19} \\

\hline
\multicolumn{15}{c}{\rule{0pt}{2.4ex}Affine-invariant depth map\rule[-1.0ex]{0pt}{0pt}}\\[2pt]
\hline
DAP~\cite{lin2025dap} & 6.80 & 22.69 & 93.92 & 98.04 & 8.98 & 35.03 & 89.68 & \textcolor{blue}{96.16} & 5.53 & 24.55 & 95.54 & 97.88 & 7.10 & 93.05  \\
UniK3D~\cite{piccinelli2025unik3d} & 9.02 & 22.72 & 92.92 & \textcolor{blue}{98.45} & \textcolor{lightgray}{8.14} & \textcolor{lightgray}{28.80} & \textcolor{lightgray}{94.34} & \textcolor{lightgray}{97.87} & 8.06 & \textcolor{blue}{19.25} & 94.02 & \textcolor{red}{99.13} & 8.41 & 93.76 \\
DA360~\cite{jiang2025depth} & \textcolor{blue}{6.55} & \textcolor{blue}{21.13} & \textcolor{blue}{94.93} & 97.59 & \textcolor{blue}{7.92} & \textcolor{blue}{32.75} & \textcolor{blue}{91.39} & 96.00 & \textcolor{blue}{4.70} & {20.29} & \textcolor{blue}{97.26} & {98.24} & \textcolor{blue}{6.39} & \textcolor{blue}{94.53} \\
{DepthMaster (Ours)} &
\textcolor{red}{4.59} & \textcolor{red}{17.59} & \textcolor{red}{96.27} & \textcolor{red}{98.56} &
\textcolor{red}{5.79} & \textcolor{red}{24.68} & \textcolor{red}{94.99} & \textcolor{red}{98.14} &
\textcolor{red}{3.36} & \textcolor{red}{15.32} & \textcolor{red}{97.74} & \textcolor{blue}{98.99} &
\textcolor{red}{4.58} & \textcolor{red}{96.33} \\

\hline
\multicolumn{15}{c}{\rule{0pt}{2.4ex}Metric depth map\rule[-1.0ex]{0pt}{0pt}} \\[2pt]
\hline
DAP~\cite{lin2025dap} & \textcolor{blue}{10.89} & \textcolor{blue}{28.45} & \textcolor{blue}{91.44} & \textcolor{blue}{98.03} & \textcolor{blue}{17.78} & \textcolor{blue}{83.40} & \textcolor{blue}{81.92} & \textcolor{blue}{93.78} & \textcolor{blue}{6.71} & \textcolor{blue}{25.12} & \textcolor{blue}{94.84} & \textcolor{blue}{97.69} & \textcolor{blue}{11.79} & \textcolor{blue}{89.40}  \\
UniK3D~\cite{piccinelli2025unik3d} & 22.14 & 45.75 & 68.79 & 97.05 & \textcolor{lightgray}{21.74} & \textcolor{lightgray}{54.51} & \textcolor{lightgray}{69.74} & \textcolor{lightgray}{96.32} & 21.69 & 39.89 & 71.48 & 95.45 & 21.86 & 70.00 \\
{DepthMaster (Ours)} &
\textcolor{red}{7.73} & \textcolor{red}{20.04} & \textcolor{red}{95.58} & \textcolor{red}{98.55} &
\textcolor{red}{9.49} & \textcolor{red}{29.85} & \textcolor{red}{91.65} & \textcolor{red}{97.65} &
\textcolor{red}{4.54} & \textcolor{red}{15.99} & \textcolor{red}{97.34} & \textcolor{red}{98.76} &
\textcolor{red}{7.25} & \textcolor{red}{94.86} \\
\bottomrule
\end{NiceTabular}}
\label{tab:main}
\vspace{-4mm}
\end{table*}

\begin{figure*}[t!]
    \centering
    \includegraphics[width=0.95\linewidth]{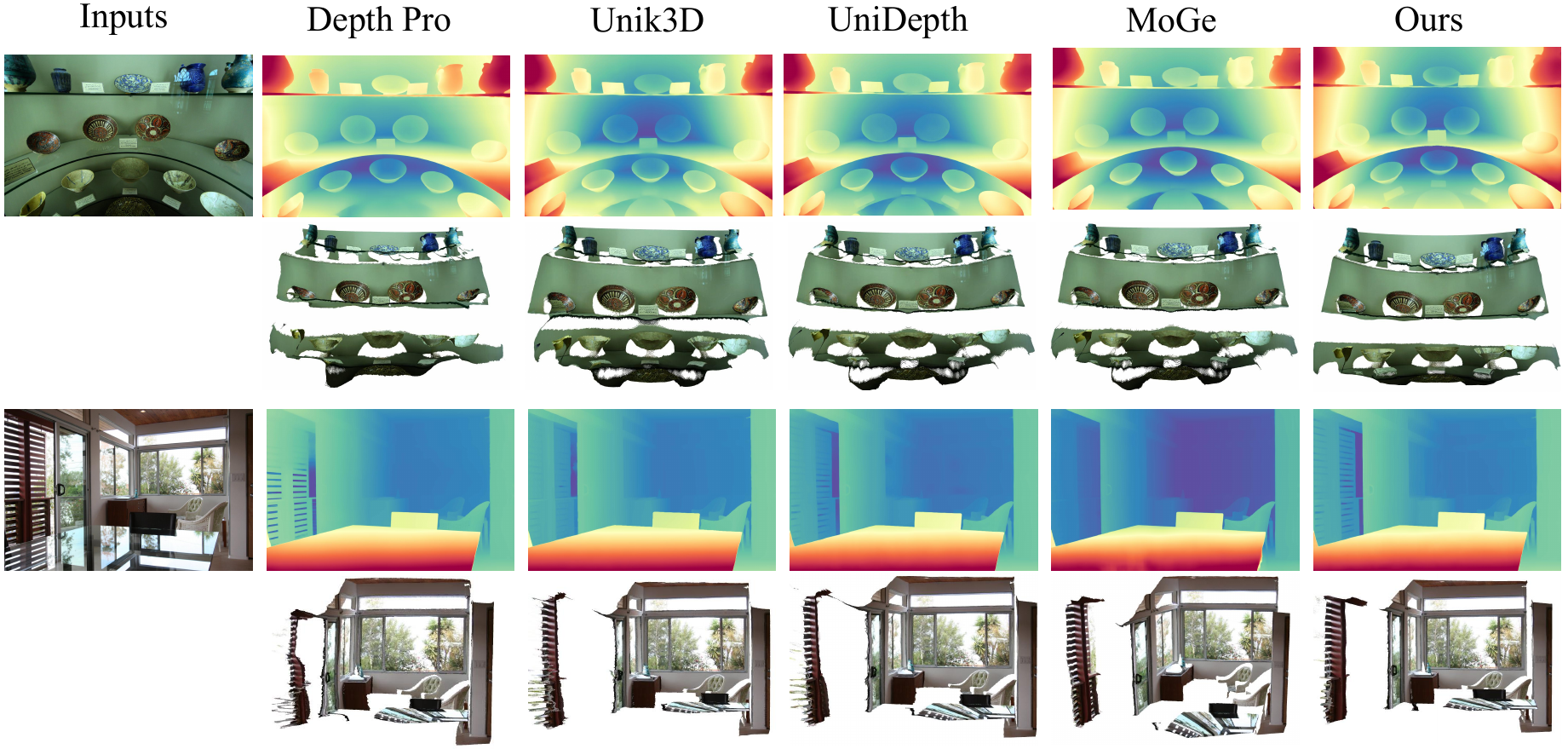}
    \vspace{-3mm}
    \caption{\textbf{Qualitative comparison on perspective images.} Compared with state-of-the-art models (Depth Pro, UniK3D, UniDepth, and MoGe), DepthMaster demonstrates a robust capability to recover fine structures, preserve sharp boundaries, and maintain accurate geometric consistency.}
    \label{fig:vis_pers}
    \vspace{-3mm}
\end{figure*}

\begin{figure*}[t!]
    \centering
    \includegraphics[width=0.95\linewidth]{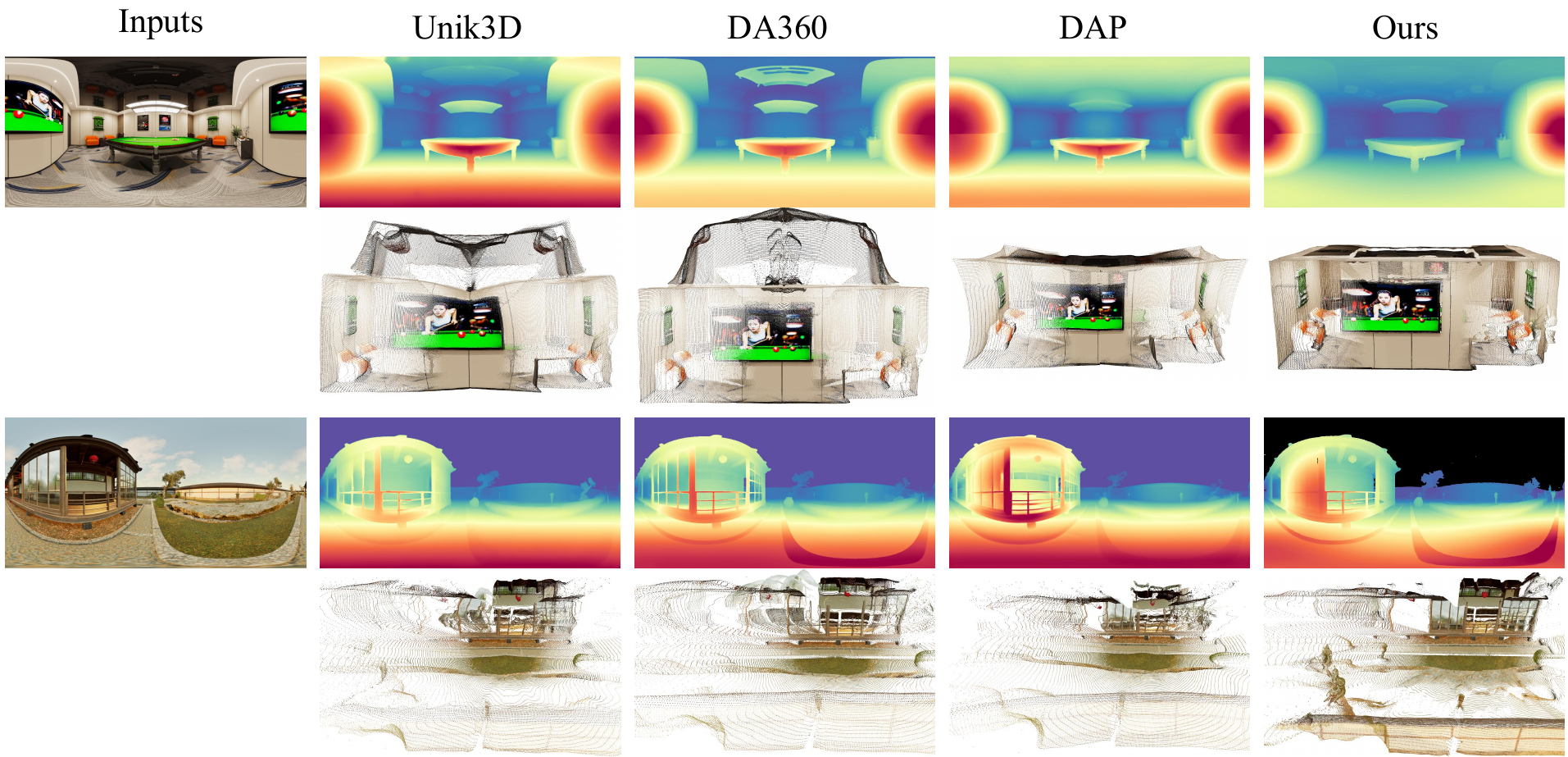} 
    \vspace{-2mm}
    \caption{\textbf{Qualitative comparison on panoramic images.} Compared with UniK3D, DA360, and DAP, our method excels at maintaining geometric accuracy and is free of the severe distortion artifacts commonly encountered in other methods.}
    \label{fig:vis_pano}
    \vspace{-5mm}
\end{figure*}

\noindent\textbf{Qualitative Comparison.}
As shown in Figure~\ref{fig:vis_pers} and Figure~\ref{fig:vis_pano}, DepthMaster generates depth maps with superior fidelity and structural consistency. 
It effectively recovers fine details and preserves sharp boundaries, whereas competing unified models like UniK3D often produce over-smoothed details and visible artifacts. 
These qualitative results align with our quantitative findings, further validating the effectiveness of our design. Due to space constraints, we provide additional qualitative comparisons in Appendix~\ref{Appendix:qualitative} to demonstrate the strong generalization ability. 
We strongly encourage readers to explore the interactive demos on our project page: \url{https://polyu-vclab.github.io/DepthMaster-page}.

\noindent\textbf{Impact of CCL Loss.}
Removing $\mathcal{L}_{\text{CCL}}$ has a negligible effect on perspective images but raises the panoramic AbsRel from 6.31 to 7.41 ($\delta_1$: 93.3\%,$\rightarrow$,92.4\%). As illustrated in Figure \ref{fig:ccl}, conspicuous geometric inconsistencies and visible seams appear in the overlapping regions. This confirms that CCL is essential for merging patch-level estimates into a globally coherent depth map on panorama inputs.

\noindent\textbf{Impact of Virtual Camera Pose.}
Omitting camera pose conditioning only mildly affects perspective performance (AbsRel 7.15,$\rightarrow$,7.47), but substantially degrades panoramic results (AbsRel 6.31,$\rightarrow$,14.36, $\delta_1$ 93.3\%,$\rightarrow$,85.8\%). Explicit pose conditioning is therefore crucial for resolving the complex spatial geometry of panoramic scenes.

\noindent\textbf{Impact of Unified Perspective Representation.}
We train a ``Naive'' baseline that removes the canonicalization module and directly feeds a mixture of perspective and ERP inputs to the network. Perspective performance degrades moderately (AbsRel 7.15,$\rightarrow$,7.74), while panoramic performance drops sharply (AbsRel 6.31,$\rightarrow$,9.45, $\delta_1$ 93.3\%,$\rightarrow$,89.7\%), confirming that a single network struggles to reconcile fundamentally different projections.

\noindent\textbf{Impact of Training Data.}
We train a ``Panoramic-Only'' variant on the single panoramic dataset while keeping the same canonicalization pipeline, so the network still sees cubemap patches. Due to limited scene diversity and the domain gap between panorama-derived patches and natural photographs, perspective performance drops severely (AbsRel 7.15,$\rightarrow$,16.34, $\delta_1$ 91.8\%,$\rightarrow$,78.1\%), and panoramic performance also degrades (AbsRel 6.31,$\rightarrow$,8.28, $\delta_1$ 93.3\%,$\rightarrow$,91.5\%). This shows that large-scale perspective data not only preserve perspective generalization but also enhance panoramic depth estimation via cross-domain prior transfer.

\noindent\textbf{Prior Transferability and Efficiency.} 
Our method surpasses the SOTA DAP method~\cite{lin2025dap} with significantly less panoramic data. It utilizes only 35K panorama training samples, while DAP requires 1.9M panoramic samples. This demonstrates the powerful prior transferability of our framework, effectively unlocking perspective priors for the data-scarce panoramic domain. Meanwhile, when benchmarking on an A100 GPU at a resolution of 1024$\times$2048, DepthMaster's inference time is 531 ms (vs. 427 ms for DAP), which remains acceptable given the substantial accuracy gains.

\begin{table*}[t!]
    \scriptsize
    \setlength{\tabcolsep}{2pt}
    \centering
    \caption{Quantitative evaluation for \emph{metric and relative geometry}, comparing specialized and unified models. 
        The \textcolor{red}{best} and \textcolor{blue}{second best} results are highlighted in \textcolor{red}{red} and \textcolor{blue}{blue}, respectively. 
        \textcolor{lightgray}{Gray numbers} denote models trained on this dataset. 
        We adopt the results reported in MoGe \cite{wang2025moge} and MoGe V2 \cite{wang2025moge2} for all methods, excluding the unified model. Unavailable/unsupported metrics are marked with `\textcolor{lightgray}{-}'.
        The complete quantitative results for each individual dataset are provided in Appendix~\ref{Appendix:full_results}.}
    \begin{tabular}{l|cc|cc|cc|cc|cc|cc|cc|cc|c}
        \specialrule{.12em}{0em}{0em}
        
        \multirow{3}{*}{\textbf{Method}} 
        & \multicolumn{8}{c|}{Point} 
        & \multicolumn{8}{c|}{Depth} 
        & \textit{Avg.}
        \\ 

        & \multicolumn{2}{c|}{\scriptsize Metric} 
        & \multicolumn{2}{c|}{\scriptsize Scale-inv.} 
        & \multicolumn{2}{c|}{\scriptsize Affine-inv.}
        & \multicolumn{2}{c|}{\scriptsize Local}
        & \multicolumn{2}{c|}{\scriptsize Metric}
        & \multicolumn{2}{c|}{\scriptsize Scale-inv.}
        & \multicolumn{2}{c|}{\scriptsize Affine-inv.}
        & \multicolumn{2}{c|}{\scriptsize Affine-inv. \tiny(disp)}
        & \\ 
        
        &\scriptsize Rel\textsuperscript{p}\scriptsize$\downarrow$
        &\scriptsize $\delta_1^\text{p}$\scriptsize$\uparrow$ 
        &\scriptsize Rel\textsuperscript{p}\scriptsize$\downarrow$
        &\scriptsize $\delta_1^\text{p}$\scriptsize$\uparrow$ 
        &\scriptsize Rel\textsuperscript{p}\scriptsize$\downarrow$
        &\scriptsize $\delta_1^\text{p}$\scriptsize$\uparrow$ 
        &\scriptsize Rel\textsuperscript{p}\scriptsize$\downarrow$
        &\scriptsize $\delta_1^\text{p}$\scriptsize$\uparrow$ 
        &\scriptsize Rel\textsuperscript{d}\scriptsize$\downarrow$
        &\scriptsize $\delta_1^\text{d}$\scriptsize$\uparrow$ 
        &\scriptsize Rel\textsuperscript{d}\scriptsize$\downarrow$
        &\scriptsize $\delta_1^\text{d}$\scriptsize$\uparrow$ 
        &\scriptsize Rel\textsuperscript{d}\scriptsize$\downarrow$
        &\scriptsize $\delta_1^\text{d}$\scriptsize$\uparrow$ 
        &\scriptsize Rel\textsuperscript{d}\scriptsize$\downarrow$
        &\scriptsize $\delta_1^\text{d}$\scriptsize$\uparrow$ 
        &\scriptsize Rk.$\downarrow$
        \\
        \hline

        ZoeDepth~\cite{bhat2023zoedepth}  
        & \textcolor{lightgray}{--} & \textcolor{lightgray}{--} 
        & \textcolor{lightgray}{--} & \textcolor{lightgray}{--} 
        & \textcolor{lightgray}{--} & \textcolor{lightgray}{--} 
        & \textcolor{lightgray}{--} & \textcolor{lightgray}{--} 
        & \textcolor{lightgray}{39.3} & \textcolor{lightgray}{49.9} 
        & \textcolor{lightgray}{12.7} & \textcolor{lightgray}{83.9} 
        & \textcolor{lightgray}{10.1} & \textcolor{lightgray}{88.5} 
        & \textcolor{lightgray}{11.1} & \textcolor{lightgray}{88.3} 
        & 10.25
        \\
        
        DA V1~\cite{yang2024depth1}  
        & \textcolor{lightgray}{--} & \textcolor{lightgray}{--} 
        & \textcolor{lightgray}{--} & \textcolor{lightgray}{--} 
        & \textcolor{lightgray}{--} & \textcolor{lightgray}{--} 
        & \textcolor{lightgray}{--} & \textcolor{lightgray}{--} 
        & \textcolor{lightgray}{31.8} & \textcolor{lightgray}{54.8} 
        & \textcolor{lightgray}{11.7} & \textcolor{lightgray}{85.8} 
        & \textcolor{lightgray}{8.76} & \textcolor{lightgray}{90.4} 
        & \textcolor{lightgray}{8.63} & \textcolor{lightgray}{92.2} 
        & 8.00
        \\
        
        DA V2~\cite{yang2024depth2}  
        & \textcolor{lightgray}{--} & \textcolor{lightgray}{--} 
        & \textcolor{lightgray}{--} & \textcolor{lightgray}{--} 
        & \textcolor{lightgray}{--} & \textcolor{lightgray}{--} 
        & \textcolor{lightgray}{--} & \textcolor{lightgray}{--} 
        & 29.9 & 56.6 
        & 10.7 & 87.6 
        & 8.48 & 90.8 
        & 8.82 & 91.6 
        & 7.13
        \\ 
        
        Metric3D V2~\cite{hu2024metric3d}
        & \textcolor{lightgray}{--} & \textcolor{lightgray}{--} 
        & \textcolor{lightgray}{--} & \textcolor{lightgray}{--} 
        & \textcolor{lightgray}{--} & \textcolor{lightgray}{--} 
        & \textcolor{lightgray}{--} & \textcolor{lightgray}{--} 
        & \textcolor{lightgray}{--} & \textcolor{lightgray}{--} 
        & \textcolor{lightgray}{7.92} & \textcolor{lightgray}{91.8} 
        & \textcolor{lightgray}{7.66} & \textcolor{lightgray}{92.9} 
        & \textcolor{lightgray}{9.51} & \textcolor{lightgray}{89.4} 
        & 5.67
        \\

        MASt3R~\cite{mast3r_eccv24}
        & 26.2 & 55.3 
        & 14.5 & 82.1 
        & 11.6 & 86.0 
        & 8.09 & 92.2 
        & 49.7 & 30.3 
        & 11.2 & 86.5 
        & 9.38 & 89.1 
        & 11.6 & 87.8 
        & 8.25
        \\
        
        UniDepth V1 ~\cite{piccinelli2024unidepth} 
        & 12.1 & 87.2 
        & 13.6 & 85.0 
        & 10.9 & 88.1 
        & 9.21 & 91.0 
        & 23.2 & 67.5 
        & 10.1 & 89.1 
        & 8.61 & 91.0 
        & 9.75 & 89.9 
        & 6.44
        \\

        Depth Pro~\cite{bochkovskii2024depth}  
        & 13.7 & 81.9 
        & 12.4 & 87.7 
        & 9.93 & 89.4 
        & 6.91 & 94.1 
        & 27.6 & 54.4 
        & 9.81 & 89.1 
        & 7.65 & 92.0 
        & 8.42 & 91.7 
        & 5.50
        \\
        
        UniDepth V2~\cite{piccinelli2025unidepthv2}  
        & 10.1 & 91.9 
        & 11.6 & 87.7 
        & 8.56 & 91.9 
        & 6.34 & 94.9 
        & 21.3 & 75.3 
        & 8.61 & 90.8 
        & 6.42 & 93.9 
        & 7.35 & 93.0 
        & 3.75
        \\

        UniK3D~\cite{piccinelli2025unik3d}
        & \textcolor{red}{8.18} & 92.3 
        & 10.7 & 88.9 
        & 8.18 & 92.3 
        & 6.25 & 94.9 
        & \textcolor{blue}{15.5} & \textcolor{red}{82.5} 
        & 8.26 & 91.1 
        & 6.50 & 93.8 
        & 7.38 & 93.0 
        & 2.75
        \\

        MoGe 2~\cite{wang2025moge2} 
        & \textcolor{blue}{8.19} & \textcolor{blue}{93.6} 
        & 10.8 & 88.5 
        & \textcolor{blue}{7.98} & 91.7 
        & \textcolor{red}{5.33} & \textcolor{blue}{95.9} 
        & 15.7 & 76.8 
        & \textcolor{blue}{7.35} & \textcolor{blue}{92.2} 
        & \textcolor{blue}{5.62} & \textcolor{blue}{94.8} 
        & \textcolor{red}{6.66} & \textcolor{red}{93.8} 
        & \textcolor{blue}{2.22}
        \\

        Ours
        & 8.27 & \textcolor{red}{93.9} 
        & \textcolor{red}{8.12} & \textcolor{red}{92.1} 
        & \textcolor{red}{6.18} & \textcolor{red}{93.9} 
        & \textcolor{blue}{5.34} & \textcolor{red}{96.0} 
        & \textcolor{red}{14.2} & \textcolor{blue}{82.3} 
        & \textcolor{red}{6.76} & \textcolor{red}{92.7} 
        & \textcolor{red}{5.18} & \textcolor{red}{94.9} 
        & \textcolor{blue}{6.80} & \textcolor{red}{93.8} 
        & \textcolor{red}{1.31}
        \\

        \specialrule{.12em}{0em}{0em}
    \end{tabular}
    \vspace{-10pt}
    \label{tab:quantitative_comparison_relative_and_metric}
\end{table*}

\begin{figure*}[t!]
    \centering
    \begin{minipage}[c]{0.47\textwidth} 
        \centering
        \captionof{table}{Ablation study on the key components in DepthMaster across different image types. }
        \label{tab:ablation}
        \resizebox{0.9\linewidth}{!}{%
            \setlength{\tabcolsep}{2pt}
            \begin{NiceTabular}{lcc|cc}
            \toprule
            \multirow{2}{*}{Model Variant} &
            \multicolumn{2}{c}{Perspective} &
            \multicolumn{2}{c}{Panoramic} \\
            \cmidrule(lr){2-3} \cmidrule(lr){4-5}
            & Rel$\downarrow$ & $\delta_1\!\uparrow$ &
            Rel$\downarrow$ & $\delta_1\!\uparrow$ \\
            \midrule
            Naive Baseline      & 7.74 & 91.0 & 9.45 & 89.7 \\
            w/o CCL             & 7.28 & 91.7 & 7.41 & 92.4 \\
            w/o Camera Pose     & 7.47 & 91.4 & 14.36 & 85.8 \\
            Panoramic-Only      & 16.34 & 78.1 & 8.28 & 91.5 \\
            DepthMaster (Full)  & \textbf{7.15} & \textbf{91.8} & \textbf{6.31} & \textbf{93.3} \\
            \bottomrule
            \end{NiceTabular}
        }
    \end{minipage}
    \hfill 
    \begin{minipage}[c]{0.48\textwidth}
        \centering
        \vspace{6mm}
        \includegraphics[width=\linewidth]{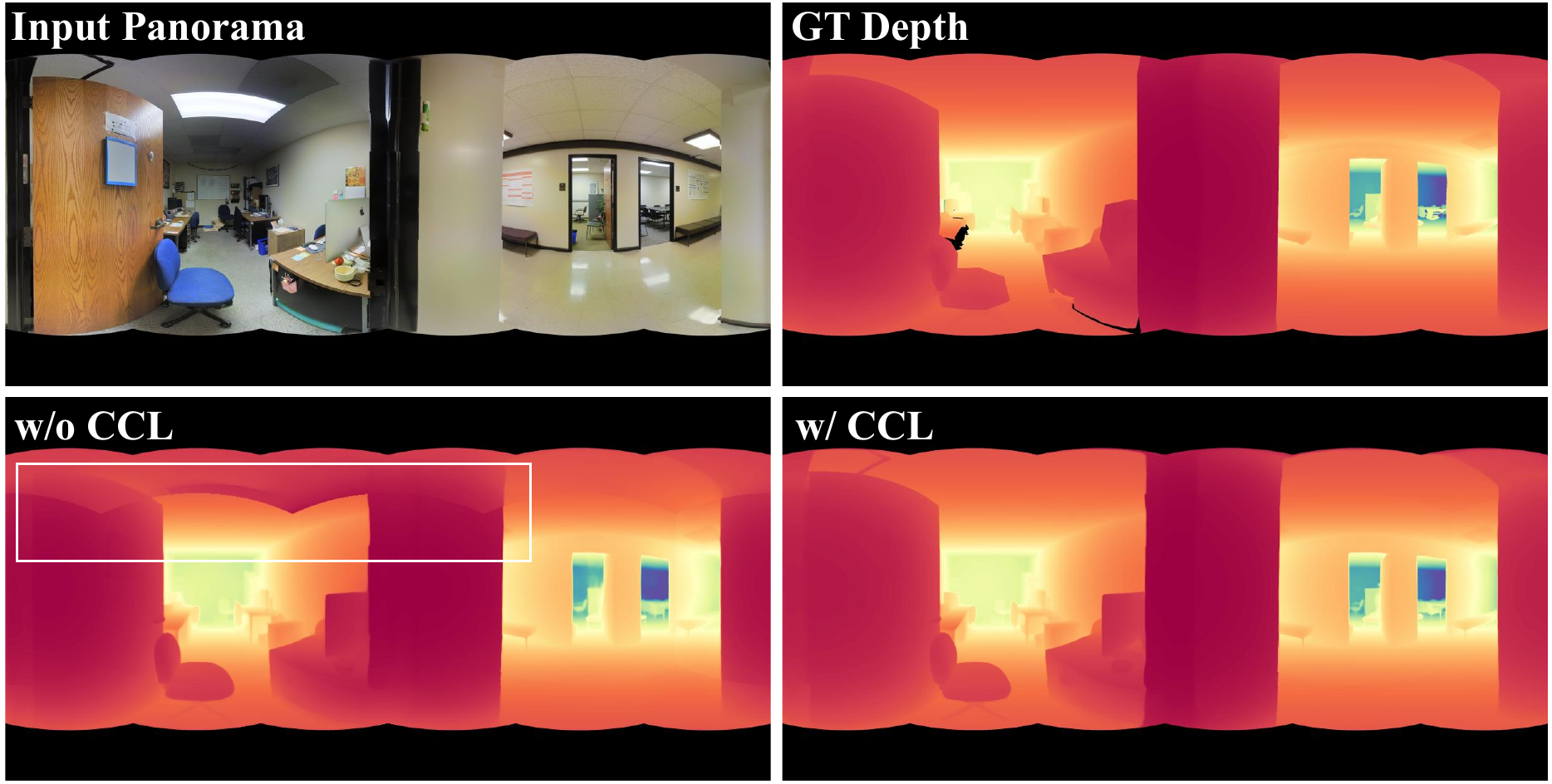}
        \vspace{-5mm}
        \caption{Impact of the proposed CCL loss on depth prediction quality.}
        \label{fig:ccl}
    \end{minipage}
    \vspace{-3mm}
\end{figure*}
\section{Conclusion}

In this paper, we introduced DepthMaster, a unified monocular depth estimation framework for both perspective and panoramic images. Our core strategy was to canonicalize any input image into standard perspective patches, thereby overcoming the challenges of geometric variance and data scarcity of panoramic cameras. A novel correspondence consistency loss was proposed to ensure geometric consistency across these patches. Extensive evaluations demonstrated that DepthMaster achieved new state-of-the-art performance on 13 diverse datasets of different camera types, surpassing existing universal and specialized methods. Our work presented a simple, powerful, and scalable paradigm for building universal geometric perception systems.

\noindent \textbf{Limitations.} While effective, DepthMaster still has some limitations. Specifically, generalization to highly abstract out-of-distribution (OOD) data remains a common challenge to it and almost all state-of-the-art models. Like existing methods, DepthMaster encounters difficulties with artistic images that deviate significantly from the training data, such as ink wash paintings (see Appendix~\ref{Appendix:limitations} for examples). In the future, we plan to address this limitation by leveraging image editing and style transfer models to incorporate more diverse training data.



{\small
\bibliography{ref}
}

\appendix

\clearpage
\setcounter{table}{0}
\setcounter{equation}{0}
\setcounter{figure}{0}
\renewcommand{\thetable}{\thesection.\arabic{table}}
\renewcommand{\theequation}{\thesection.\arabic{equation}}
\renewcommand{\thefigure}{\thesection.\arabic{figure}}
\setcounter{table}{0}
\renewcommand{\thetable}{A\arabic{table}}
\setcounter{figure}{0}
\renewcommand{\thefigure}{A\arabic{figure}}
\setcounter{equation}{0}
\numberwithin{equation}{section}

\begin{center}
    {\LARGE\bfseries Appendix\par}
\end{center}
\vspace{0.5em}

In this appendix, we provide the following materials:
\begin{itemize}[leftmargin=4em,labelwidth=1.5em,labelsep=0.7em,align=left,itemsep=0.3em]
    \item[\textbf{A}] Implementation details, covering training configurations and the unified evaluation protocols adopted for both perspective and panoramic datasets (referring to Sec.~\ref{Appendix:implementation}).
    \item[\textbf{B}] Complete the evaluation results on individual perspective benchmarks, offering a granular view beyond the aggregated metrics reported in the main paper (referring to Sec.~\ref{Appendix:full_results}).
    \item[\textbf{C}] Detailed descriptions of the training and validation datasets used in our experiments, including their scale, modality, and sampling ratios (referring to Sec.~\ref{Appendix:data}).
    \item[\textbf{D}] More qualitative comparisons with state-of-the-art methods across diverse indoor, outdoor, and panoramic scenes (referring to Sec.~\ref{Appendix:qualitative}).
    \item[\textbf{E}] Discussion of current limitations of DepthMaster and potential future research directions (referring to Sec.~\ref{Appendix:limitations}).
\end{itemize}

\section{Implementation Details}\label{Appendix:implementation}

\subsection{Training Details}\label{Appendix:Details}

We base our model architecture on DINOv2~\cite{oquab2023dinov2} and initialize with its pre-trained weights. For optimization, we employ the AdamW optimizer with a cosine decay learning rate schedule, setting the learning rate to $5 \times 10^{-5}$. During the training phase, we uniformly resize the input images to a width of 518 pixels, followed by cropping to achieve an aspect ratio uniformly distributed in the range of [1.0, 2.0]. The model is trained for 200K iterations on 32 NVIDIA H100 GPUs with a total batch size of 256. To enhance the model's robustness, we employ a series of data augmentation techniques, including color jittering, Gaussian blur, and JPEG compression artifacts. Given that our training data includes only one panorama dataset, we assign it a fixed sampling ratio of 20\% to ensure its adequate representation during training.

\textbf{Loss Weights.} The total training objective is a weighted combination of the point map loss, the surface normal loss, the metric scale loss, and the validity mask loss. Specifically, we set the weights of the point map loss and the surface normal loss to $1.0$, and the weights of the metric scale loss and the validity mask loss to $0.1$, i.e.,
\begin{equation}
    \mathcal{L} = 1.0 \cdot \mathcal{L}_{\text{point}} + 1.0 \cdot \mathcal{L}_{\text{normal}} + 0.1 \cdot \mathcal{L}_{\text{scale}} + 0.1 \cdot \mathcal{L}_{\text{mask}}.
\end{equation}
This configuration balances the contributions from the dense geometric supervision (point map and surface normal) and the auxiliary metric and validity supervision, and is kept fixed throughout all training stages.

\subsection{Evaluation Protocol}

\textbf{Relative Geometry}. To ensure a rigorous and fair evaluation of relative geometry, we adopt the alignment strategy introduced in MoGe~\cite{wang2025moge}. Before calculating the errors, the predictions are aligned with the ground truth data using optimal scale and shift (where applicable) for each individual sample. The specific alignment formulations are detailed as follows:

\begin{itemize} [leftmargin=*]
    \item \textit{Scale-invariant point map.} We align the predicted point map $\hat{\mathbf{p}}_i$ with the ground truth $\mathbf{p}_i$ by optimizing a global scale factor $a^*$. The objective is to minimize the depth-weighted $L_1$ distance:
    \begin{equation}
        a^* = \mathop{\arg\!\min}_a \sum_{i\in\mathcal M} \frac{1}{z_i}\|a\hat {\mathbf{p}}_i-\mathbf{p}_i\|_1.
    \end{equation}

    \item \textit{Affine-invariant point map.} When both scale and translation are involved, we compute the optimal scale $a^*$ and shift $\mathbf{b}^*$ to minimize the alignment error:
    \begin{equation}
        (a^*,\mathbf{b}^*) = \mathop{\arg\!\min}_{a,\mathbf{b}} \sum_{i\in\mathcal M} \frac{1}{z_i}\|a\hat{\mathbf{p}}_i+\mathbf{b}-\mathbf{p}_i\|_1.
    \end{equation}
    
    \item \textit{Scale-invariant depth map.} For depth map evaluation, the optimal scaling factor $a^*$ is derived by minimizing the depth-weighted absolute difference:
    \begin{equation}
        a^* = \mathop{\arg\!\min}_a \sum_{i\in\mathcal M} \frac{1}{z_i}|a\hat{z}_i-z_i|.
    \end{equation}

    \item \textit{Affine-invariant depth map.} The optimal scale $a^*$ and offset $b^*$ for an affine transformation of the depth map are solved via:
    \begin{equation}
        (a^*, b^*) = \mathop{\arg\!\min}_{a, b} \sum_{i\in\mathcal M} \frac{1}{z_i}|a\hat{z}_i+b-z_i|.
    \end{equation}

    \item \textit{Affine-invariant disparity map.} In line with standard practices for disparity-based alignment~\cite{ranftl2020midas}, we perform a least-squares optimization in the disparity space. The scale $a^*$ and shift $b^*$ are determined by:
    \begin{equation}
        (a^*, b^*) = \mathop{\arg\!\min}_{a, b} \sum_{i\in\mathcal M}(a\hat{d}_i+b-d_i)^2,
    \end{equation}
    where $\hat{d}_i$ represents the predicted disparity and $d_i = 1/z_i$ represents the ground truth disparity. To avoid invalid depth values resulting from zero or negative disparities, the transformed disparity is lower-bounded by $1/z_\text{max}$ (the inverse of the maximum valid depth). The final aligned depth $\hat{z}_i^*$ is thus obtained by:
    \begin{equation}
        \hat{z}_i^* := \frac{1}{\max(a^* \hat{d}_i + b^*, 1/z_\text{max})}.
    \end{equation}
\end{itemize}

\textbf{Metric Geometry}. The following geometry related metrics are used in the evaluation:

\begin{itemize} [leftmargin=*]
    \item \textit{Metric depth.} For metric depth evaluation, we directly assess the model's raw output against the ground truth without any alignment or value clipping, except when a method explicitly requires specific post-processing as part of its standard inference pipeline.

    \item \textit{Metric point map.} To evaluate metric point maps, the predictions are only aligned via a global translation vector $\mathbf{b}^*$, which is calculated to minimize the depth-weighted spatial error:
    \begin{equation}
        \mathbf{b}^* = \mathop{\arg\!\min}_{\mathbf{b}} \sum_{i\in\mathcal M} \frac{1}{z_i}\|\hat{\mathbf{p}}_i+\mathbf{b}-\mathbf{p}_i\|_1.
    \end{equation}
        
\end{itemize}

\subsection{Evaluation Details}\label{Appendix:eval}
The methods presented in Table 1 and Table 2 of the main paper often employ disparate evaluation protocols, making a direct comparison challenging. To ensure a fair and consistent assessment across our diverse test set of perspective and panoramic data, we adopt a unified protocol for each camera format. Specifically, one standardized protocol is applied to all perspective datasets, and a separate one is used for all panoramic datasets.

For perspective datasets, as described in the main text, we adhere to the established MoGe protocol~\cite{wang2025moge}. At inference time, images from various datasets with differing original resolutions are resized to a height of 518 pixels, while the width is scaled proportionally to maintain the aspect ratio. The resolution of the ground truth (GT) used for evaluation is consistent with the MoGe protocol, and our model's predictions are subsequently resized via interpolation to match the GT resolution before evaluation.

For panoramic datasets, we adhere to the DA$^{2}$ protocol~\cite{li2025depth}. At inference time, each panoramic (equirectangular) image is projected into six 518$\times$518 perspective patches. The final 512$\times$1024 depth prediction map is then generated by back-projecting the results from the cube faces to the equirectangular format. The GT resolution for evaluation aligns with the DA$^{2}$ protocol, and our predicted depths are interpolated to match this resolution.

\section{Complete Evaluation on Individual Datasets} \label{Appendix:full_results}

In the main text of the paper, we have provided detailed evaluation results for panoramic datasets. However, due to space constraints, the performance on perspective image benchmarks was primarily reported as average metrics across multiple datasets to provide a macroscopic view of our model's capabilities. Evaluating a universal monocular depth estimation model necessitates a granular analysis across diverse scenes. Therefore, in this supplementary material, we provide the complete, unaggregated evaluation results specifically for perspective scenes. 

Table~\ref{tab:full_comparison} lists the performance of DepthMaster alongside state-of-the-art baselines on each individual perspective dataset used in our evaluation suite (such as NYUv2, KITTI, ETH3D, etc.). By presenting these granular results, we offer a more transparent and comprehensive understanding of the model's robustness and specific strengths across different domains in perspective depth estimation.


\begin{table*}[!htbp]
    \scriptsize
    \setlength{\tabcolsep}{1.2pt}
    \centering
        \caption{Evaluation results of our method and competing methods on each dataset.}
    \label{tab:full_comparison}
    \vspace{-5pt}
    \begin{tabular}{l|cc|cc|cc|cc|cc|cc|cc|cc|cc|cc|cc}
        \specialrule{.12em}{0em}{0em}
        
        \multirow{2}{*}{\textbf{Method}} 
        & \multicolumn{2}{c|}{NYUv2} 
        & \multicolumn{2}{c|}{KITTI} 
        & \multicolumn{2}{c|}{ETH3D} 
        & \multicolumn{2}{c|}{iBims-1} 
        & \multicolumn{2}{c|}{GSO} 
        & \multicolumn{2}{c|}{Sintel} 
        & \multicolumn{2}{c|}{DDAD} 
        & \multicolumn{2}{c|}{DIODE} 
        & \multicolumn{2}{c|}{Spring}
        & \multicolumn{2}{c|}{HAMMER}
        & \multicolumn{2}{c}{\textit{Avg.}} \\
        
        &\scriptsize Rel\scriptsize$\downarrow$
        &\scriptsize $\delta_1$\scriptsize$\uparrow$ 
        &\scriptsize Rel\scriptsize$\downarrow$
        &\scriptsize $\delta_1$\scriptsize$\uparrow$ 
        &\scriptsize Rel\scriptsize$\downarrow$
        &\scriptsize $\delta_1$\scriptsize$\uparrow$ 
        &\scriptsize Rel\scriptsize$\downarrow$
        &\scriptsize $\delta_1$\scriptsize$\uparrow$ 
        &\scriptsize Rel\scriptsize$\downarrow$
        &\scriptsize $\delta_1$\scriptsize$\uparrow$ 
        &\scriptsize Rel\scriptsize$\downarrow$
        &\scriptsize $\delta_1$\scriptsize$\uparrow$ 
        &\scriptsize Re\scriptsize$\downarrow$
        &\scriptsize $\delta_1$\scriptsize$\uparrow$ 
        &\scriptsize Rel\scriptsize$\downarrow$
        &\scriptsize $\delta_1$\scriptsize$\uparrow$ 
        &\scriptsize Rel\scriptsize$\downarrow$
        &\scriptsize $\delta_1$\scriptsize$\uparrow$ 
        &\scriptsize Rel\scriptsize$\downarrow$
        &\scriptsize $\delta_1$\scriptsize$\uparrow$ 
        &\scriptsize Rel\scriptsize$\downarrow$
        &\scriptsize $\delta_1$\scriptsize$\uparrow$ \\
        
        \hline
        \multicolumn{23}{c}{Metric point map} \\
        \hline

UniDepth V1 & 4.80 & 98.3 & 4.52 & 98.5 & 22.4 & 63.1 & 10.8 & 92.8 & - & - & - & - & 11.4 & 89.5 & 12.8 & 88.9 & - & - & 18.0 & 79.5 & 12.1 & 87.2 \\

UniDepth V2 & 4.83 & 98.0 & 5.88 & 97.5 & 9.46 & 95.0 & 5.23 & 97.9 & - & - & - & - & 13.3 & 90.3 & 17.0 & 80.8 & - & - & 15.0 & 83.9 & 10.1 & 91.9 \\

UniK3D & 4.51 & 97.9 & 5.41 & 97.7 & 7.79 & 97.0 & 4.56 & 98.3 & - & - & - & - & 14.8 & 89.9 & 8.23 & 93.9 & - & - & 14.3 & 85.0 & 8.18 & 92.3 \\

Depth Pro & 6.13 & 97.3 & 11.1 & 85.3 & 21.2 & 64.9 & 6.89 & 96.9 & - & - & - & - & 22.6 & 61.3 & 13.5 & 81.8 & - & - & 14.5 & 86.0 & 13.7 & 81.9 \\

MoGe 2 & 4.44 & 98.3 & 7.44 & 94.4 & 7.19 & 97.7 & 5.63 & 97.4 & - & - & - & - & 11.4 & 87.9 & 7.85 & 92.3 & - & - & 13.4 & 87.0 & 8.19 & 93.6 \\

\emph{DepthMaster (Ours)} & 5.19 & 97.8 & 8.89 & 94.9 & 8.78 & 93.7 & 5.49 & 97.2 & - & - & - & - & 11.4 & 86.7 & 8.35 & 92.6 & - & - & 9.80 & 95.1 & 8.27 & 93.9 \\
        \hline
        \multicolumn{23}{c}{Metric depth map (wo/ GT intrinsics)} \\
        \hline

DA V1 & 10.5 & 94.9 & 11.6 & 94.5 & 40.2 & 24.0 & 12.9 & 81.8 & - & - & - & - & 34.5 & 44.7 & 58.0 & 16.2 & - & - & 54.8 & 27.3 & 31.8 & 54.8 \\

DA V2 & 16.4 & 80.9 & 10.6 & 88.6 & 36.1 & 36.3 & 11.1 & 91.7 & - & - & - & - & 41.7 & 37.5 & 41.2 & 22.1 & - & - & 52.1 & 38.9 & 29.9 & 56.6 \\

UniDepth V1 & 7.59 & 97.6 & 4.69 & 98.4 & 56.9 & 14.9 & 23.8 & 57.6 & - & - & - & - & 13.8 & 85.1 & 17.1 & 71.9 & - & - & 38.2 & 46.7 & 23.2 & 67.5 \\

UniDepth V2 & 10.6 & 92.8 & 8.58 & 95.4 & 20.7 & 69.5 & 9.52 & 93.2 & - & - & - & - & 18.4 & 77.6 & 43.0 & 51.8 & - & - & 38.2 & 46.8 & 21.3 & 75.3 \\

UniK3D & 8.87 & 94.2 & 7.97 & 96.4 & 15.7 & 82.2 & 9.06 & 94.3 & - & - & - & - & 20.7 & 73.8 & 17.2 & 76.4 & - & - & 28.9 & 60.2 & 15.5 & 82.5 \\

Depth Pro & 10.7 & 91.9 & 23.5 & 38.3 & 38.5 & 32.8 & 15.9 & 81.5 & - & - & - & - & 33.4 & 35.3 & 31.9 & 37.7 & - & - & 39.1 & 63.0 & 27.6 & 54.4 \\

MoGe 2 & 7.33 & 96.1 & 18.1 & 62.9 & 10.4 & 90.8 & 13.6 & 83.0 & - & - & - & - & 15.8 & 73.0 & 17.5 & 66.4 & - & - & 26.9 & 65.6 & 15.7 & 76.8 \\
                
\emph{DepthMaster (Ours)} & 8.78 & 95.4 & 15.0 & 83.5 & 15.5 & 78.6 & 12.5 & 89.0 & - & - & - & - & 14.7 & 80.2 & 17.4 & 70.2 & - & - & 15.7 & 79.1 & 14.2 & 82.3 \\

        \hline
        \multicolumn{23}{c}{Scale-invariant point map} \\
        \hline

UniDepth V1 & 5.33 & 98.4 & 5.96 & 98.5 & 18.5 & 77.6 & 5.29 & 97.4 & 6.58 & 99.6 & 33.0 & 48.9 & 11.4 & 90.2 & 12.3 & 91.0 & 33.1 & 49.8 & 4.83 & 98.5 & 13.6 & 85.0 \\

UniDepth V2 & 5.59 & 98.3 & 5.41 & 98.0 & 6.58 & 97.2 & 5.56 & 98.1 & 4.53 & 99.7 & 27.2 & 56.3 & 13.4 & 91.2 & 12.0 & 93.4 & 31.9 & 46.0 & 4.20 & 99.2 & 11.6 & 87.7 \\

UniK3D & 4.73 & 98.2 & 5.08 & 98.1 & 5.69 & 98.1 & 4.42 & 98.1 & 3.55 & 99.9 & 25.6 & 58.5 & 15.3 & 89.9 & 8.70 & 94.3 & 29.2 & 54.8 & 5.06 & 99.5 & 10.7 & 88.9 \\

Depth Pro & 5.04 & 97.7 & 10.6 & 95.1 & 11.2 & 92.0 & 5.84 & 97.1 & 4.94 & 99.8 & 26.9 & 63.9 & 15.8 & 81.0 & 8.52 & 91.6 & 28.1 & 60.5 & 6.82 & 98.7 & 12.4 & 87.7 \\

MoGe & 4.86 & 98.4 & 5.47 & 97.4 & 4.58 & 98.9 & 4.63 & 97.1 & 2.58 & 100 & 22.3 & 69.5 & 12.3 & 90.3 & 6.58 & 94.5 & 4.84 & 96.4 & 6.45 & 98.1 & 7.46 & 94.1 \\

MoGe 2 & 3.94 & 98.3 & 8.27 & 97.5 & 5.45 & 98.6 & 5.34 & 98.3 & 2.55 & 100 & 23.1 & 66.8 & 11.0 & 90.7 & 8.42 & 93.7 & 31.1 & 42.4 & 8.77 & 98.4 & 10.8 & 88.5 \\
    
\emph{DepthMaster (Ours)} & 4.02 & 97.9 & 5.60 & 96.8 & 4.56 & 98.8 & 4.34 & 98.1 & 2.84 & 100 & 18.8 & 72.2 & 10.5 & 89.0 & 5.97 & 95.2 & 19.7 & 73.3 & 4.85 & 99.6 & 8.12 & 92.1 \\

        \hline
        \multicolumn{23}{c}{Affine-invariant point map} \\
        \hline

UniDepth V1 & 3.93 & 98.4 & 4.29 & 98.6 & 12.2 & 89.6 & 4.65 & 98.0 & 2.99 & 99.8 & 28.5 & 58.4 & 10.3 & 90.5 & 8.56 & 90.9 & 29.6 & 58.5 & 4.15 & 98.7 & 10.9 & 88.1 \\

UniDepth V2 & 3.66 & 98.4 & 4.75 & 98.0 & 4.35 & 98.4 & 4.05 & 98.1 & 2.91 & 99.9 & 17.9 & 76.5 & 12.0 & 90.8 & 7.45 & 92.4 & 25.1 & 66.9 & 3.45 & 99.4 & 8.56 & 91.9 \\

UniK3D & 3.41 & 98.3 & 4.47 & 98.0 & 3.85 & 98.7 & 3.37 & 98.3 & 1.91 & 100 & 18.1 & 76.9 & 13.7 & 89.2 & 5.71 & 94.8 & 23.8 & 69.2 & 3.44 & 99.5 & 8.18 & 92.3 \\

Depth Pro & 4.36 & 97.9 & 9.15 & 90.7 & 7.73 & 94.0 & 4.34 & 97.4 & 3.16 & 99.7 & 19.6 & 74.5 & 14.4 & 81.2 & 6.28 & 93.7 & 25.0 & 66.0 & 5.31 & 98.8 & 9.93 & 89.4 \\

MoGe & 3.68 & 98.3 & 4.86 & 97.2 & 3.57 & 99.0 & 3.61 & 97.3 & 1.14 & 100 & 16.8 & 77.8 & 10.5 & 91.4 & 4.37 & 96.4 & 4.45 & 96.4 & 3.88 & 98.1 & 5.69 & 95.2 \\

MoGe 2 & 3.33 & 98.4 & 6.47 & 96.4 & 3.89 & 98.7 & 3.65 & 98.5 & 1.16 & 100 & 17.4 & 77.0 & 10.1 & 90.3 & 5.13 & 94.9 & 24.5 & 63.7 & 4.19 & 99.1 & 7.98 & 91.7 \\

\emph{DepthMaster (Ours)} & 3.54 & 98.0 & 5.02 & 97.2 & 3.18 & 99.1 & 3.13 & 97.9 & 1.33 & 100 & 15.3 & 79.4 & 9.50 & 89.8 & 4.07 & 96.2 & 14.9 & 82.0 & 1.79 & 99.7 & 6.18 & 93.9 \\

        \hline
        \multicolumn{23}{c}{Local point map} \\
        \hline
        
MASt3R & - & - & - & - & 5.54 & 95.3 & 6.19 & 95.0 & - & - & 11.4 & 87.9 & 8.58 & 91.8 & 8.75 & 90.9 & - & - & - & - & 8.09 & 92.2 \\

UniDepth V1 & - & - & - & - & 8.61 & 92.6 & 5.92 & 96.0 & - & - & 13.4 & 84.3 & 8.18 & 92.0 & 9.95 & 90.0 & - & - & - & - & 9.21 & 91.0 \\

UniDepth V2 & - & - & - & - & 3.99 & 97.4 & 4.02 & 97.3 & - & - & 9.35 & 92.2 & 8.18 & 92.4 & 6.15 & 95.3 & - & - & - & - & 6.34 & 94.9 \\

UniK3D & - & - & - & - & 3.84 & 97.4 & 3.92 & 97.2 & - & - & 9.12 & 92.4 & 8.55 & 92.3 & 5.81 & 95.3 & - & - & - & - & 6.25 & 94.9 \\

Depth Pro & - & - & - & - & 4.76 & 96.9 & 4.11 & 97.5 & - & - & 10.8 & 89.5 & 8.08 & 92.4 & 6.80 & 94.1 & - & - & - & - & 6.91 & 94.1 \\

MoGe & - & - & - & - & 3.21 & 98.1 & 4.16 & 96.8 & - & - & 8.63 & 92.7 & 6.74 & 94.3 & 4.78 & 96.3 & - & - & - & - & 5.50 & 95.6 \\

MoGe 2 & - & - & - & - & 3.27 & 98.2 & 3.61 & 97.7 & - & - & 8.13 & 93.2 & 6.57 & 94.3 & 5.09 & 96.1 & - & - & - & - & 5.33 & 95.9 \\

\emph{DepthMaster (Ours)} & - & - & - & - & 3.51 & 98.1 & 3.87 & 97.5 & - & - & 7.28 & 94.4 & 6.69 & 94.2 & 5.30 & 95.8 & - & - & - & - & 5.34 & 96.0 \\

        \hline
        \multicolumn{23}{c}{Scale-invariant depth map} \\
        \hline

MASt3R & 5.37 & 96.0 & 6.24 & 94.5 & 5.68 & 95.5 & 5.58 & 95.2 & 3.72 & 99.1 & 26.3 & 63.7 & 13.5 & 81.5 & 8.37 & 89.4 & 32.2 & 53.5 & 5.50 & 96.5 & 11.2 & 86.5 \\

DA V1 & 4.77 & 97.5 & 5.61 & 95.6 & 9.41 & 88.9 & 5.53 & 95.8 & 5.49 & 99.3 & 28.3 & 56.7 & 13.2 & 81.5 & 10.3 & 87.5 & 27.3 & 59.1 & 6.88 & 96.4 & 11.7 & 85.8 \\

DA V2 & 5.03 & 97.3 & 7.23 & 93.7 & 6.12 & 95.5 & 4.32 & 97.9 & 4.38 & 99.3 & 23.0 & 65.2 & 14.7 & 78.0 & 7.95 & 90.0 & 28.0 & 61.1 & 5.92 & 97.7 & 10.7 & 87.6 \\

Metric3D V2 & 4.69 & 97.4 & 4.00 & 98.5 & 3.84 & 98.5 & 4.23 & 97.7 & 2.46 & 99.9 & 20.7 & 69.8 & 7.41 & 94.6 & 3.29 & 98.4 & 24.4 & 64.4 & 4.19 & 99.1 & 7.92 & 91.8 \\

UniDepth V1 & 3.86 & 98.4 & 3.73 & 98.6 & 5.67 & 97.0 & 4.79 & 97.4 & 4.18 & 99.7 & 28.3 & 58.8 & 10.1 & 90.5 & 6.83 & 92.8 & 29.2 & 59.3 & 4.19 & 98.4 & 10.1 & 89.1 \\

UniDepth V2 & 3.65 & 98.4 & 4.24 & 98.0 & 3.23 & 98.9 & 3.45 & 98.1 & 3.16 & 99.7 & 23.1 & 65.3 & 11.0 & 91.5 & 5.92 & 94.1 & 24.9 & 65.1 & 3.48 & 99.1 & 8.61 & 90.8 \\

UniK3D & 3.36 & 98.4 & 4.08 & 98.1 & 3.33 & 98.7 & 3.42 & 98.0 & 2.21 & 99.9 & 21.4 & 67.5 & 11.8 & 91.4 & 5.22 & 94.7 & 24.6 & 65.1 & 3.17 & 99.5 & 8.26 & 91.1 \\

Depth Pro & 4.42 & 97.6 & 5.47 & 96.2 & 7.54 & 94.1 & 4.13 & 97.4 & 2.18 & 99.9 & 23.9 & 68.7 & 14.0 & 82.0 & 7.05 & 92.0 & 25.1 & 63.8 & 4.36 & 98.9 & 9.81 & 89.1 \\

MoGe & 3.44 & 98.4 & 4.25 & 97.8 & 3.36 & 98.9 & 3.46 & 97.0 & 1.47 & 100 & 19.3 & 73.4 & 9.17 & 90.5 & 4.89 & 94.7 & 4.63 & 96.4 & 3.77 & 98.1 & 5.77 & 94.5 \\

MoGe 2 & 3.44 & 98.2 & 4.11 & 98.0 & 3.55 & 98.7 & 3.16 & 98.2 & 1.49 & 100 & 19.6 & 71.6 & 8.91 & 91.2 & 5.30 & 94.6 & 20.0 & 72.4 & 3.96 & 99.2 & 7.35 & 92.2 \\

\emph{DepthMaster (Ours)} & 3.60 & 97.9 & 5.03 & 97.0 & 2.86 & 98.8 & 3.20 & 98.1 & 1.63 & 100 & 16.1 & 77.1 & 9.44 & 88.9 & 4.80 & 95.2 & 17.7 & 74.8 & 3.19 & 99.7 & 6.76 & 92.7 \\

        \hline
        \multicolumn{23}{c}{Affine-invariant depth} \\
        \hline

MASt3R & 4.67 & 96.7 & 5.79 & 95.1 & 4.64 & 97.0 & 4.62 & 95.6 & 2.85 & 99.4 & 21.3 & 70.3 & 12.5 & 83.4 & 5.79 & 94.1 & 27.4 & 62.8 & 4.21 & 96.8 & 9.38 & 89.1 \\

DA V1 & 3.82 & 98.3 & 5.04 & 96.4 & 6.23 & 95.2 & 4.23 & 97.3 & 1.98 & 100 & 20.1 & 71.8 & 11.3 & 86.1 & 6.75 & 92.6 & 22.4 & 68.9 & 5.77 & 97.3 & 8.76 & 90.4 \\

DA V2 & 4.16 & 97.9 & 6.77 & 94.3 & 4.63 & 97.2 & 3.44 & 98.3 & 1.44 & 100 & 17.1 & 76.6 & 13.4 & 81.8 & 5.41 & 94.6 & 23.7 & 68.7 & 4.73 & 98.9 & 8.48 & 90.8 \\

Metric3D V2 & 3.94 & 97.6 & 3.50 & 98.4 & 3.24 & 99.0 & 3.28 & 98.3 & 2.10 & 99.4 & 26.6 & 71.7 & 7.15 & 94.8 & 2.75 & 98.7 & 21.0 & 72.5 & 3.02 & 99.0 & 7.66 & 92.9 \\

UniDepth V1 & 3.40 & 98.6 & 3.55 & 98.7 & 4.92 & 97.5 & 3.76 & 98.2 & 2.48 & 99.9 & 24.9 & 64.1 & 9.46 & 90.8 & 4.90 & 96.2 & 25.2 & 67.3 & 3.55 & 98.9 & 8.61 & 91.0 \\

UniDepth V2 & 2.96 & 98.6 & 3.85 & 98.1 & 2.95 & 98.5 & 2.64 & 98.4 & 1.37 & 100 & 13.3 & 83.2 & 10.5 & 90.9 & 4.05 & 96.5 & 20.1 & 75.4 & 2.48 & 99.6 & 6.42 & 93.9 \\

UniK3D & 2.81 & 98.6 & 3.77 & 98.1 & 2.96 & 98.7 & 2.56 & 98.4 & 1.19 & 100 & 13.7 & 83.0 & 11.7 & 89.5 & 3.62 & 97.0 & 20.3 & 75.3 & 2.43 & 99.6 & 6.50 & 93.8 \\

Depth Pro & 3.67 & 98.2 & 5.12 & 96.8 & 4.97 & 96.4 & 3.23 & 98.3 & 1.46 & 100 & 15.8 & 80.1 & 12.6 & 84.1 & 4.66 & 95.6 & 21.7 & 70.5 & 3.30 & 99.6 & 7.65 & 92.0 \\

MoGe & 2.92 & 98.6 & 3.94 & 98.0 & 2.69 & 99.2 & 2.74 & 97.9 & 0.94 & 100 & 13.0 & 83.2 & 8.40 & 92.1 & 3.16 & 97.5 & 4.34 & 96.4 & 3.00 & 98.3 & 4.51 & 96.1 \\

MoGe 2 & 2.89 & 98.6 & 3.75 & 98.1 & 2.80 & 99.1 & 2.36 & 98.8 & 0.94 & 100 & 13.3 & 82.5 & 8.26 & 92.5 & 3.14 & 97.4 & 15.9 & 81.2 & 2.85 & 99.3 & 5.62 & 94.8 \\

\emph{DepthMaster (Ours)} & 3.15 & 98.3 & 4.43 & 97.2 & 2.37 & 99.3 & 2.51 & 98.3 & 1.09 & 100 & 12.5 & 83.1 & 8.67 & 90.5 & 2.86 & 97.6 & 13.1 & 84.5 & 1.10 & 99.7 & 5.18 & 94.9 \\

        \hline
        \multicolumn{23}{c}{Affine-invariant disparity} \\
        \hline

DA V1 & 4.20 & 98.4 & 5.40 & 97.0 & 4.68 & 98.2 & 4.18 & 97.6 & 1.54 & 100 & 20.2 & 77.6 & 12.7 & 86.9 & 5.69 & 95.7 & 22.2 & 72.5 & 5.56 & 98.0 & 8.63 & 92.2 \\

DA V2 & 4.14 & 98.3 & 5.61 & 96.7 & 4.71 & 97.9 & 3.47 & 98.5 & 1.24 & 100 & 21.4 & 72.8 & 13.1 & 86.4 & 5.29 & 96.1 & 24.3 & 70.6 & 4.97 & 99.1 & 8.82 & 91.6 \\

Metric3D V2 & 13.4 & 81.5 & 3.76 & 98.2 & 4.30 & 97.7 & 8.55 & 92.3 & 1.80 & 100 & 21.8 & 72.4 & 7.35 & 94.1 & 7.70 & 90.2 & 23.3 & 68.1 & 3.17 & 99.2 & 9.51 & 89.4 \\

MASt3R & 5.07 & 96.8 & 5.93 & 95.5 & 5.25 & 96.4 & 5.39 & 95.7 & 2.98 & 99.7 & 30.2 & 65.1 & 13.0 & 83.6 & 6.41 & 94.3 & 37.3 & 53.2 & 4.41 & 97.2 & 11.6 & 87.8 \\

UniDepth V1 & 3.78 & 98.7 & 3.64 & 98.7 & 5.34 & 97.2 & 4.06 & 98.1 & 2.56 & 99.9 & 28.6 & 60.7 & 9.94 & 89.1 & 5.95 & 95.5 & 30.0 & 61.6 & 3.64 & 99.1 & 9.75 & 89.9 \\

UniDepth V2 & 3.38 & 98.7 & 3.99 & 98.0 & 2.97 & 99.0 & 3.15 & 98.3 & 1.30 & 100 & 17.2 & 79.9 & 10.2 & 90.2 & 4.43 & 96.4 & 24.4 & 69.6 & 2.51 & 99.6 & 7.35 & 93.0 \\

UniK3D & 3.24 & 98.6 & 3.88 & 98.1 & 2.89 & 99.0 & 3.03 & 98.4 & 1.16 & 100 & 17.0 & 80.0 & 10.5 & 90.1 & 3.95 & 97.0 & 25.8 & 68.9 & 2.47 & 99.6 & 7.38 & 93.0 \\

Depth Pro & 4.21 & 98.1 & 5.10 & 97.0 & 4.94 & 96.7 & 3.74 & 98.2 & 1.49 & 100 & 17.4 & 79.1 & 11.7 & 87.1 & 4.84 & 96.4 & 27.5 & 64.5 & 3.31 & 99.6 & 8.42 & 91.7 \\

MoGe & 3.38 & 98.6 & 4.05 & 98.1 & 3.11 & 98.9 & 3.23 & 98.0 & 0.96 & 100 & 18.4 & 79.5 & 8.99 & 91.5 & 3.98 & 97.2 & 6.43 & 93.7 & 3.30 & 98.5 & 5.58 & 95.4 \\

MoGe 2 & 3.35 & 98.6 & 3.92 & 98.1 & 3.21 & 98.9 & 2.85 & 98.7 & 0.96 & 100 & 18.0 & 78.7 & 8.69 & 92.1 & 4.03 & 97.2 & 18.7 & 76.6 & 2.90 & 99.5 & 6.66 & 93.8 \\

\emph{DepthMaster (Ours)} & 3.63 & 98.4 & 4.71 & 97.0 & 2.95 & 98.8 & 3.05 & 98.2 & 1.11 & 100 & 19.2 & 80.2 & 9.24 & 90.1 & 3.67 & 97.3 & 18.9 & 77.9 & 1.47 & 99.7 & 6.80 & 93.8 \\

        \specialrule{.12em}{0em}{0em}
    \end{tabular}
\end{table*}

\section{Training and Validation Datasets}\label{Appendix:data}
This section details the datasets used for training and evaluating DepthMaster. Our methodology aims to achieve robust zero-shot monocular depth estimation for both perspective and panoramic images. This is achieved by training the model on a large-scale, mixed-data regime and subsequently validating its generalization capabilities on a diverse suite of unseen evaluation benchmarks. These benchmarks were intentionally selected to encompass both perspective and panoramic formats, as well as diverse data distributions that span indoor and outdoor environments and comprise both real-world and synthetic scenes, to rigorously test the model's versatility. The specifics of our training and validation sets are detailed below.

\subsection{Training Datasets}
\Cref{tab:supp:train_ds} summarizes the composition of our training set, including each dataset's domain, frame count, type, and the final calculated sampling weight.
Our training dataset is a curated aggregation of 17 publicly available datasets from diverse domains, totaling approximately 4.0 million frames. This collection comprises one panoramic dataset, Structured3D~\cite{Structured3D}, and 14 perspective datasets. To balance these heterogeneous sources during training, we employ a principled sampling strategy. Given that our training data includes only one panoramic dataset, we assign it a fixed sampling ratio of 20\% to ensure its adequate representation.

The remaining 80\% of the sampling budget is allocated to the 16 perspective datasets, including Hypersim~\cite{roberts2021hypersim}, A2D2~\cite{geyer2020a2d2}, MVS-Synth~\cite{huang2018mvsynth}, GTA-SfM~\cite{Wang2019gtasfm}, UrbanSyn~\cite{gómez2023urbansyn}, TartanAir~\cite{tartanair2020iros}, IRS~\cite{wang2019IRS}, ARKitScenes~\cite{dehghan2021arkitscenes}, MegaDepth~\cite{MegaDepthLi18}, BlendedMVS~\cite{yao2020blendedmvs}, KenBurns~\cite{niklaus2019kenburns}, Waymo~\cite{sun2020waymo}, Argoverse2~\cite{Argoverse2}, Taskonomy-tiny~\cite{zamir2018taskonomy}, CARD \cite{elazab2026card}, and PointOdyssey \cite{zheng2023pointodyssey}. The sampling ratio for each dataset is given in Table~\Cref{tab:supp:train_ds}.

\subsection{Validation Datasets}
To assess the zero-shot generalization capability of DepthMaster, we evaluate it on a comprehensive benchmark of 13 distinct datasets, as summarized in \Cref{tab:eval_datasets_reformatted}. These datasets were not seen during training and were intentionally selected, including both perspective and panoramic benchmarks, allowing for a thorough evaluation of the model's flexibility. The evaluation sets are categorized as follows:
\begin{itemize}

\item \textbf{Perspective Datasets:} This category comprises ten datasets that cover a wide variety of scenes. It includes indoor environments (NYUv2~\cite{Silberman2012nyuv2}, iBims-1~\cite{ibim1_1}, HAMMER~\cite{jung2023hammer}), outdoor driving and in-the-wild scenarios (KITTI~\cite{Uhrig2017kitti}, DDAD~\cite{ddadpacking}, Spring~\cite{Mehl2023Spring}), mixed indoor/outdoor scenes (ETH3D~\cite{Schops2019ETH3D}, DIODE~\cite{diode_dataset}), object-centric views (GSO~\cite{downs2022googlescannedobjects}), and synthetic sequences (Sintel~\cite{Butler2012sintel}).

\item \textbf{Panoramic Datasets:} This category assesses the model's performance on full 360-degree equirectangular images. It consists of three large-scale indoor datasets: the real-world Stanford2D3D~\cite{stanford2d3ds} and Matterport3D~\cite{chang2017matterport3d}, and the synthetic PanoSUNCG~\cite{PanoSUNCG}. Due to the lack of established outdoor panoramic benchmarks, we provide additional qualitative comparisons for outdoor scenes in Figure~\ref{fig:pano_vis_outdoor} and on our interactive project page.

\end{itemize}
This diverse suite of evaluation benchmarks provides a rigorous testbed for validating the robustness and versatility of our model across different camera intrinsics and real-world domains.

\begin{table}[t!]
  \centering
  \caption{Training and Evaluation Datasets. Left: Training datasets. Right: Evaluation datasets.}
  \label{tab:datasets}
  \begin{minipage}[t]{0.48\textwidth}
    \centering
    \footnotesize
    \setlength{\tabcolsep}{3pt}
    \renewcommand{\arraystretch}{1.05}
    \captionsetup{skip=3pt}
    \vspace{-1mm}
    \subcaption{\textbf{Training Datasets.} SfM: Structure-from-Motion; MVS: Multi-View Stereo.}
    \label{tab:supp:train_ds}
    \resizebox{\linewidth}{!}{%
    \begin{tabular}{l c c l c}
        \specialrule{0.12em}{0em}{0em}
        Name & Domain & \#Frames & Acquisition & Weight \\
        \hline
        Hypersim & Indoor & $75$K & Synthetic & 5.12\%\\
        A2D2 & Outdoor/Driving & $196$K & LiDAR & 0.82\%\\
        MVS-Synth & Outdoor/Driving & $12$K & Synthetic & 1.23\%\\
        GTA-SfM & Outdoor/In-the-wild & $19$K & Synthetic & 2.86\% \\
        UrbanSyn & Outdoor/Driving & $7$K & Synthetic & 2.15\%\\
        TartanAir & In-the-wild & $306$K & Synthetic & 5.12\%\\
        IRS & Indoor & $101$K & Synthetic & 5.73\%\\
        ARKitScenes & Indoor & $449$K & SfM/MVS & 6.80\%\\
        MegaDepth & Outdoor/In-the-wild & $92$K & SfM/MVS & 5.73\%\\
        BlendedMVS & In-the-wild & $115$K & SfM/MVS & 12.27\%\\
        KenBurns & In-the-wild & $76$K & Synthetic & 1.64\%\\
        Waymo & Outdoor/Driving & $788$K & LiDAR & 5.00\%\\
        Argoverse2 & Outdoor/Driving & $1.1$M & LiDAR & 5.00\%\\
        Taskonomy-tiny & Indoor & $300$K & SfM/MVS & 10.42\%\\
        CARD & Outdoor/Driving & $79$K & LiDAR & 4.11\%\\
        PointOdyssey & Outdoor & $237$K  & Synthetic & 6.00\%\\
        \hline
        Structured3D & Indoor & $35$K & Synthetic & 20.00\%\\
        \specialrule{0.12em}{0em}{0em}
    \end{tabular}%
    }
  \end{minipage}
  \hspace{0.6em}
  \begin{minipage}[t]{0.45\textwidth}
    \centering
    \footnotesize
    \setlength{\tabcolsep}{3pt}
    \renewcommand{\arraystretch}{1.05}
    \captionsetup{skip=3pt}
    \subcaption{\textbf{Evaluation Datasets.} Grouped by image type: perspective (top), panoramic datasets (bottom).}
    \label{tab:eval_datasets_reformatted}
    \resizebox{\linewidth}{!}{%
    \begin{tabular}{l c c l}
        \specialrule{0.12em}{0em}{0em}
        Name & Domain & \#Frames & Acquisition \\
        \hline
        NYUv2 & Indoor & 654 & RGB-D  \\
        KITTI & Outdoor/Driving & 652 & LiDAR \\
        ETH3D & Indoor/Outdoor & 454 & SfM/MVS \\
        iBims-1 & Indoor & 100 & RGB-D  \\
        GSO & Object-centric & 1,030 & Synthetic \\
        Sintel & Synthetic/Various & 1,064 & Synthetic \\
        DDAD & Outdoor/Driving & 1,000 & LiDAR \\
        DIODE & Indoor/Outdoor & 771 & LiDAR \\
        Spring & Outdoor/In-the-wild & 1,000 & Synthetic \\
        HAMMER & Indoor & 775 & Synthetic \\
        \hline
        Stanford2D3D & Indoor & 1413  & SfM/MVS \\
        Matterport3D & Indoor & 2024  & SfM/MVS \\
        PanoSUNCG & Indoor & 3944 & Synthetic \\
        \specialrule{0.12em}{0em}{0em}
    \end{tabular}%
    }
  \end{minipage}
\end{table}

\subsection{Real Data Refinement}\label{Appendix:data_refinement}

Real-world training data often suffer from noise and incompleteness, which hamper the model's ability to accurately reconstruct fine-grained structures. Previous works~\cite{yang2024depth2,wang2025moge2} have also noted this issue and validated its impact. However, they did not open-source their data processing methods. Therefore, we implement our own strategies: depth enhancement and flying pixel filtering. The primary challenges in real-world data include blurred object boundaries and severe flying pixels at depth discontinuities. To address these issues, we introduce two key data refinement strategies: depth enhancement and flying pixel filtering.

\textbf{Depth Enhancement}.
To resolve the issue of blurred boundaries in real-world ground truth, we enhance the data using predictions from a model trained on synthetic datasets, which naturally possess perfectly sharp edges. Specifically, after obtaining the high-quality relative depth prediction $z^{\text{pred}}$ from the synthetic-trained model, we apply a robust alignment pipeline to recover the metric scale. We first employ RANSAC-based linear regression separately on near and far regions to establish a global scale $s$ and shift $t$, yielding an initial metric depth $z^{\text{init}} = s \cdot z^{\text{pred}} + t$. To address local inconsistencies, we then optimize a learnable local scale map $S \in \mathbb{R}^{H \times W}$ to refine the depth as $z^{\text{opt}} = S \odot z^{\text{init}}$. This is guided by an As-Rigid-As-Possible (ARAP) smoothness constraint and a point cloud $L_1$ loss against the noisy ground truth $z^{\text{gt}}$:
\begin{equation}
    \mathcal{L} = \sum_{p \in \mathcal{V}} \left\| S(p)\mathbf{P}^{\text{init}}(p) - \mathbf{P}^{\text{gt}}(p) \right\|_1 + \lambda \sum_{p \in \mathcal{V}} \left| S(p) - \frac{1}{|\mathcal{N}(p)|} \sum_{q \in \mathcal{N}(p)} S(q) \right|,
\end{equation}
where $\mathbf{P}(p)$ denotes the 3D coordinate unprojected from pixel $p$, $\mathcal{V}$ is the set of valid pixels, $\mathcal{N}(p)$ represents the local spatial neighborhood, and $\lambda$ is the smoothness weight. Finally, we use edge-aware Poisson blending to seamlessly integrate the refined metric depth while preserving the sharp discontinuities of the relative prediction. As illustrated in Figure~\ref{fig:depth_enhancement}, this enhancement process yields depth maps with significantly sharper edges and fine-grained structures, while maintaining robust scale alignment.

\textbf{Flying Pixel Filtering}.
To address the issue of flying pixels commonly encountered in real-world datasets, we apply an edge-based filtering technique. Specifically, we detect depth discontinuities by computing the relative depth difference between adjacent pixels. A pixel $p$ is classified as an unreliable edge region and subsequently masked out if it satisfies the following condition:
\begin{equation}
    \max_{q \in \mathcal{N}_{4}(p)} \frac{|z(p) - z(q)|}{\min(z(p), z(q))} > \tau,
\end{equation}
where $\mathcal{N}_{4}(p)$ denotes the 4-connected neighborhood of $p$, $z(p)$ is the depth value, and $\tau$ is a predefined relative tolerance threshold. This geometric strategy effectively eliminates the flying pixels commonly produced by LiDAR or SfM around object boundaries. Figure~\ref{fig:flying_pixel_filtering} demonstrates the depth maps and the corresponding 3D point clouds before and after filtering, showing a substantial reduction in flying pixels and cleaner geometric boundaries.

\begin{figure*}[t!]
    \centering
    \includegraphics[width=\linewidth]{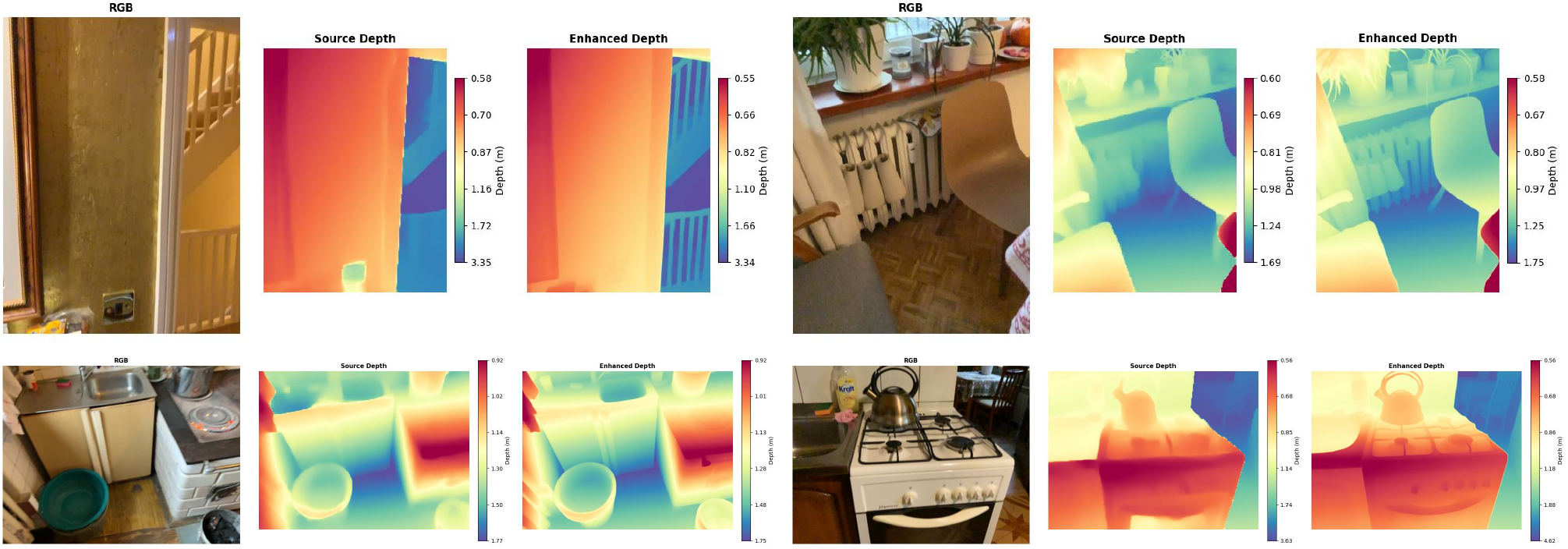}
    \caption{\textbf{Depth Enhancement on Real Data.} We enhance real-world depth maps using predictions from a synthetic-trained model followed by a scale alignment algorithm. The enhanced depth maps show significantly sharpened edges while preserving accurate scale alignment.}
    \label{fig:depth_enhancement}
\end{figure*}

\begin{figure*}[t!]
    \centering
    \includegraphics[width=\linewidth]{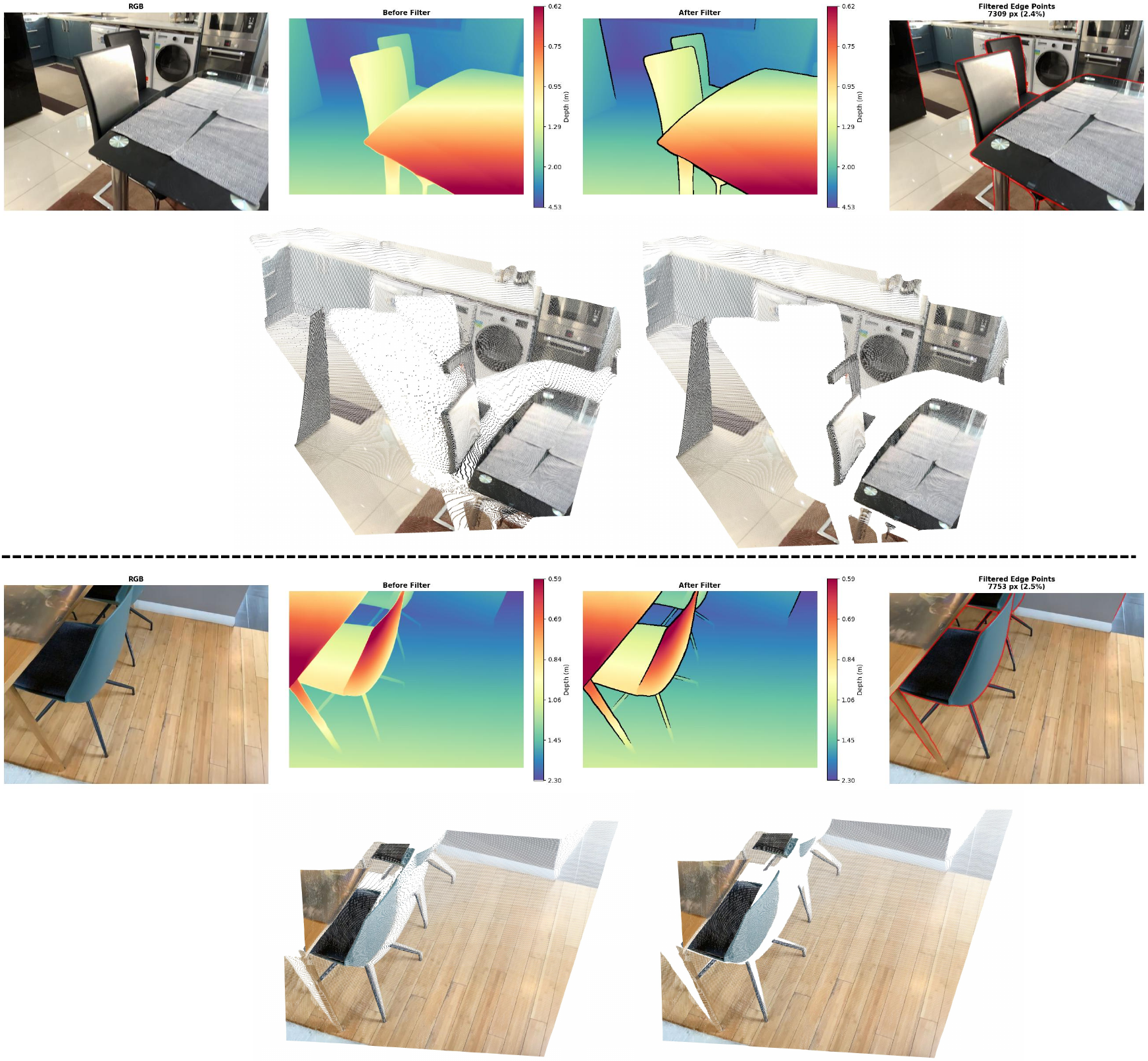}
    \caption{\textbf{Flying Pixel Filtering on Real Datasets.} We apply edge-based filtering to eliminate flying pixels at depth discontinuities. The visualizations of depth maps and 3D point clouds before and after filtering demonstrate a significant reduction in boundary artifacts.}
    \label{fig:flying_pixel_filtering}
\end{figure*}

\subsection{Training Data Scale Comparison}\label{Appendix:data_scale}

To further highlight the data efficiency of DepthMaster, we compare the training data scale of our method with representative state-of-the-art approaches in two complementary settings, as summarized in \Cref{tab:data_scale}. On the perspective side (\Cref{tab:data_scale_pers}), we report the total number of training frames used by each method for general monocular depth estimation. On the panoramic side (\Cref{tab:data_scale_pano}), we report the number of panoramic frames used for 360$^{\circ}$ depth estimation.

DepthMaster achieves superior perspective accuracy with a smaller training corpus. As shown in \Cref{tab:data_scale_pers}, our model uses only $\sim$4.0M frames, $2$--$17\times$ fewer than existing baselines (DepthAnything~V2~\cite{yang2024depth2}, Metric3D~V2~\cite{hu2024metric3d}, UniDepth~V2~\cite{piccinelli2024unidepth}, UniK3D~\cite{piccinelli2025unik3d}, MoGe~\cite{wang2025moge}, Depth Pro~\cite{Bochkovskii2024}). In particular, we deliberately exclude several large-scale real-world datasets adopted by MoGe, e.g., the Indoor subset of Taskonomy~\cite{zamir2018taskonomy} ($\sim$3.6M frames), yet our method still outperforms it on perspective benchmarks. This indicates that once the training set covers sufficiently diverse scenes, further scaling mainly adds annotation noise and scene redundancy, while our curated compact corpus is already sufficient for high-quality depth learning.

The advantage is more pronounced in the data-scarce panoramic domain, which is the central claim of DepthMaster. From \Cref{tab:data_scale_pano}, we can see that our model uses only $\sim$35K panoramic frames from a single dataset (Structured3D~\cite{Structured3D}), yet still surpasses all panorama-oriented baselines: $4\times$ fewer than PanDA~\cite{cao2025panda}, $14\times$ fewer than DA360~\cite{jiang2025depth}, $17\times$ fewer than DA$^{2}$~\cite{li2025depth}, and over $54\times$ fewer than the SOTA metric panoramic method DAP~\cite{lin2025dap} ($\sim$1.9M). A particularly informative case is UniK3D~\cite{piccinelli2025unik3d}: under the same DINOv2 ViT-Large backbone and a comparable amount of panoramic data ($\sim$39K), its panoramic accuracy still lags substantially behind ours (Tab.~\ref{tab:main} in the main paper). Therefore, the panoramic gap is not determined by the amount of 360$^{\circ}$ data, but by how it is used.

We attribute this to the prior transferability unlocked by our framework. Instead of collecting massive 360$^{\circ}$ supervision to fit the distorted ERP representation, DepthMaster reuses the high-quality perspective priors from its backbone and curated perspective data, and transfers them to 360$^{\circ}$ inputs via our canonicalization design. A small panoramic set then suffices to adapt the model to the ERP distribution, while most geometric knowledge is inherited from the perspective side. This explains two patterns in \Cref{tab:data_scale_pano}: (i) methods that purely scale up panoramic data (DA360, DA$^{2}$, DAP) exhibit clearly diminishing returns; and (ii) DepthMaster matches or exceeds their accuracy with one to two orders of magnitude less panoramic data, providing strong evidence that perspective-to-panoramic prior transfer is more effective than brute-force scaling.

\begin{table}[t!]
  \centering
  \caption{\textbf{Training data scale comparison.} All listed methods adopt DINOv2 as the backbone. Left: perspective-domain comparison (total training frames). Right: panoramic-domain comparison (panoramic training frames).}
  \label{tab:data_scale}
  \begin{minipage}[t]{0.48\textwidth}
    \centering
    \footnotesize
    \setlength{\tabcolsep}{4pt}
    \renewcommand{\arraystretch}{1.1}
    \captionsetup{skip=3pt}
    \subcaption{\textbf{Perspective-domain comparison.}}
    \label{tab:data_scale_pers}
    \resizebox{\linewidth}{!}{%
    \begin{tabular}{l c c}
        \specialrule{0.12em}{0em}{0em}
        Method & Backbone & \#Training Frames \\
        \hline
        DepthAnything~V2~\cite{yang2024depth2} & DINOv2 ViT-Large& $\sim$62M  \\
        Depth Pro~\cite{Bochkovskii2024}  & DINOv2 ViT-Large & $\sim 6$M \\
         Metric3D V2~\cite{hu2024metric3d} & DINOv2 ViT-Large & $\sim$$16$M \\
         UniDepth V2~\cite{piccinelli2024unidepth} & DINOv2 ViT-Large & $\sim$$16$M \\
        MoGe~\cite{wang2025moge} & DINOv2 ViT-Large & $\sim$9M  \\
        UniK3D~\cite{piccinelli2025unik3d} & DINOv2 ViT-Large & $\sim$10M \\
        \hline
        \textbf{DepthMaster (Ours)} & DINOv2 ViT-Large & \textbf{$\sim$4.0M} \\
        \specialrule{0.12em}{0em}{0em}
    \end{tabular}%
    }
  \end{minipage}
  \hspace{0.6em}
  \begin{minipage}[t]{0.48\textwidth}
    \centering
    \footnotesize
    \setlength{\tabcolsep}{4pt}
    \renewcommand{\arraystretch}{1.1}
    \captionsetup{skip=3pt}
    \subcaption{\textbf{Panoramic-domain comparison.}}
    \label{tab:data_scale_pano}
    \resizebox{\linewidth}{!}{%
    \begin{tabular}{l c c}
        \specialrule{0.12em}{0em}{0em}
        Method & Backbone & \#Panoramic Frames \\
        \hline
        UniK3D~\cite{piccinelli2025unik3d} & DINOv2 ViT-Large & $\sim$39K \\
        PanDA~\cite{cao2025panda} & DINOv2 ViT-Large & $\sim$141K \\
        DA360~\cite{jiang2025depth} & DINOv2 ViT-Large & $\sim$500K \\
        DA$^{2}$~\cite{li2025depth} & DINOv2 ViT-Large & $\sim$606K \\
        DAP~\cite{lin2025dap} & DINOv3 ViT-Large & $\sim$1.9M \\
        \hline
        \textbf{DepthMaster (Ours)} & DINOv2 ViT-Large & \textbf{$\sim$35K} \\
        \specialrule{0.12em}{0em}{0em}
    \end{tabular}%
    }
  \end{minipage}
\end{table}

\section{More Qualitative Comparisons}\label{Appendix:qualitative}

In this section, we present extensive qualitative comparison results to visually demonstrate the superior performance of DepthMaster on the universal monocular depth estimation task. As discussed in the main text, our model excels at generating depth maps characterized by high fidelity and structural consistency. To comprehensively evaluate its capabilities, we conduct a side-by-side comparison of DepthMaster with representative models, including UniK3D, DA360, and DAP for panoramic scenes, as well as other state-of-the-art methods for perspective scenes. These comparisons are designed to highlight our model's advantages in recovering fine-grained details, preserving sharp boundaries, and handling complex geometric structures.

\textbf{Panoramic Indoor Scenes}. As shown in Figure~\ref{fig:pano_vis_indoor}, we provide a detailed comparison between our method and competing methods (UniK3D, DA360, DAP) on indoor panoramic scenes. The visual results include both depth maps and the corresponding 3D point clouds. DepthMaster demonstrates outstanding performance in preserving precise geometric structures. In contrast, the comparison methods frequently suffer from geometric inaccuracies, such as distorted walls and inconsistent structural layouts.

\textbf{Panoramic Outdoor Scenes}. Figure~\ref{fig:pano_vis_outdoor} presents a detailed comparison of our method against UniK3D, DA360, and DAP on outdoor panoramic scenes. The depth maps clearly show that our method produces much sharper edges. Furthermore, the point cloud visualizations demonstrate that DepthMaster preserves more fine details and maintains significantly better geometric structures across various scene styles. Additionally, our method features the unique capability to predict a sky mask, which the other methods cannot achieve. For point cloud visualization, we uniformly apply our predicted sky mask across all methods to mask out the sky regions.

\textbf{Perspective Scenes}. As illustrated in Figure~\ref{fig:vis_pers_suppl}, we additionally compare our method with state-of-the-art models, including Depth Pro, UniK3D, UniDepth, and MoGe on perspective images. The visualizations, which consist of both depth maps and 3D perspective views, demonstrate that DepthMaster possesses a robust capability to preserve fine-grained details and accurate geometric structures. While baseline methods often struggle with structural consistency, our approach produces highly reliable geometric predictions.

\begin{figure*}[t!]
    \centering
    \includegraphics[width=\linewidth]{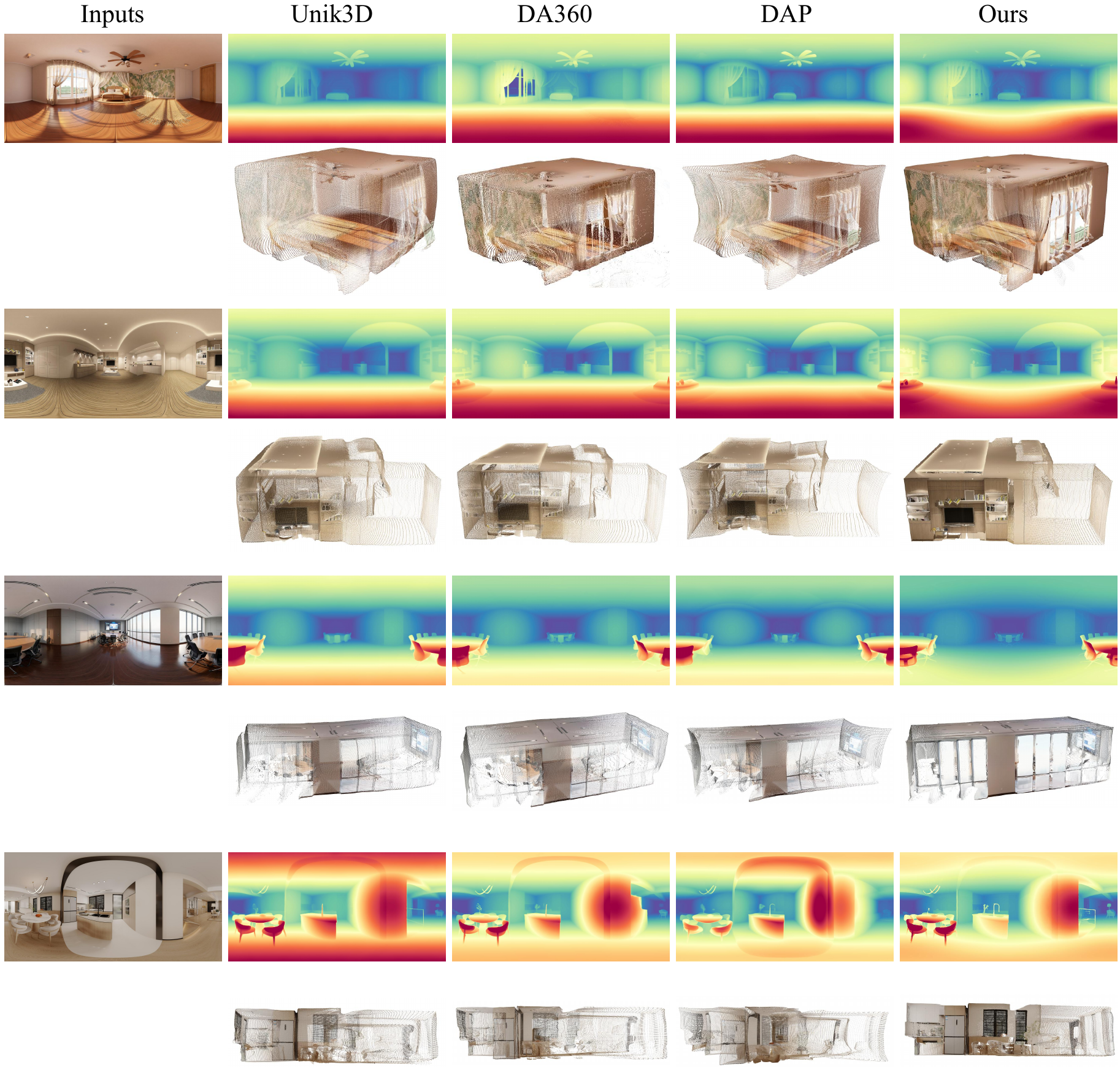}
    \caption{\textbf{Qualitative comparison on panoramic indoor scenes.} Compared with UniK3D, DA360, and DAP, DepthMaster achieves superior geometric accuracy and avoids issues such as wall distortions commonly seen in the baselines.}
    \label{fig:pano_vis_indoor}
\end{figure*}

\begin{figure*}[t!]
    \centering
    \includegraphics[width=\linewidth]{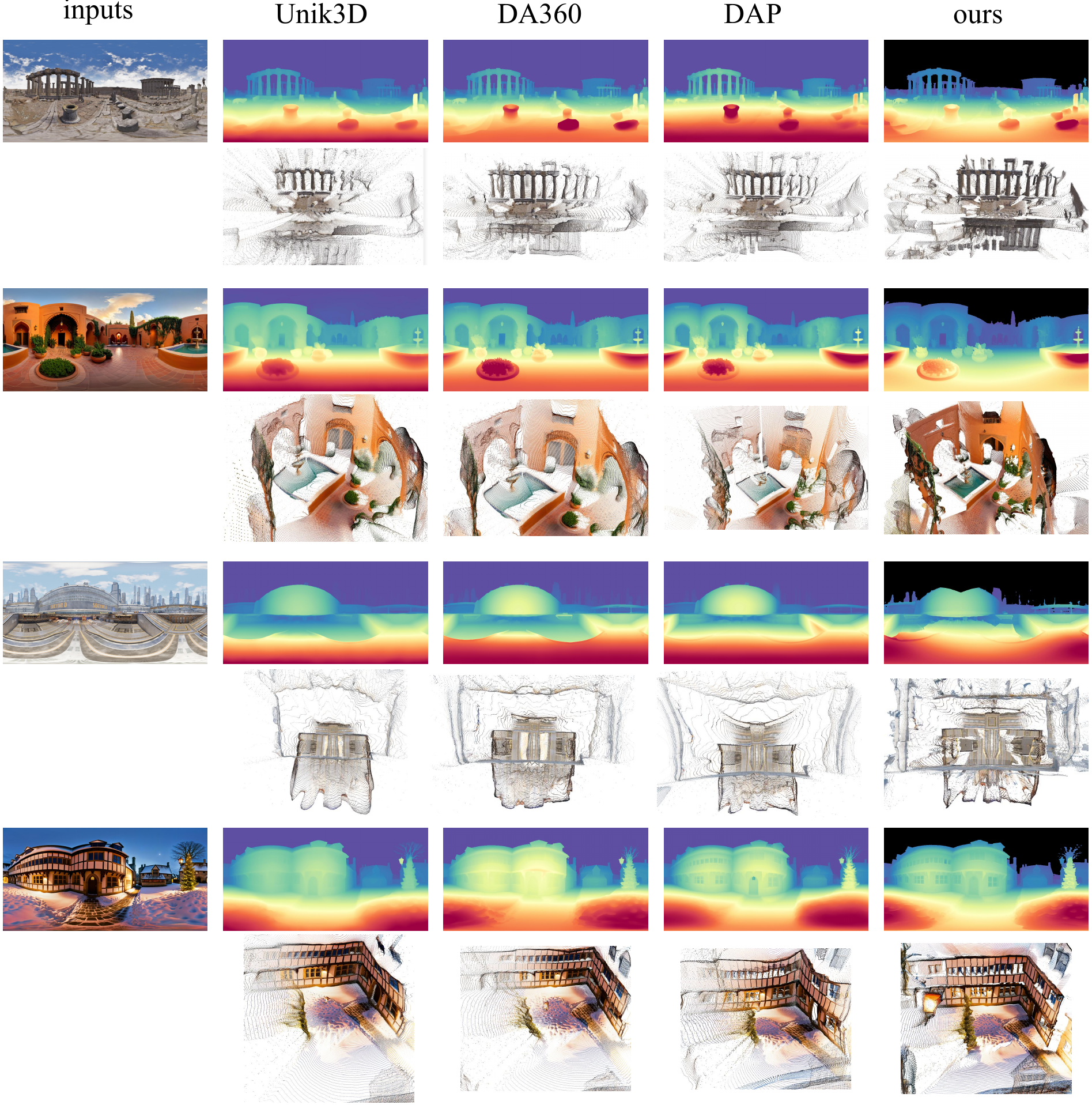}
    \caption{\textbf{Qualitative comparison on panoramic outdoor scenes.} DepthMaster yields sharper boundaries and better preserves geometric structures across different styles compared to UniK3D, DA360, and DAP. Note that all point clouds are visualized using the sky mask predicted by our method, as the baselines lack this capability.}
    \label{fig:pano_vis_outdoor}
\end{figure*}

\begin{figure*}[t!]
    \centering
    \includegraphics[width=\linewidth]{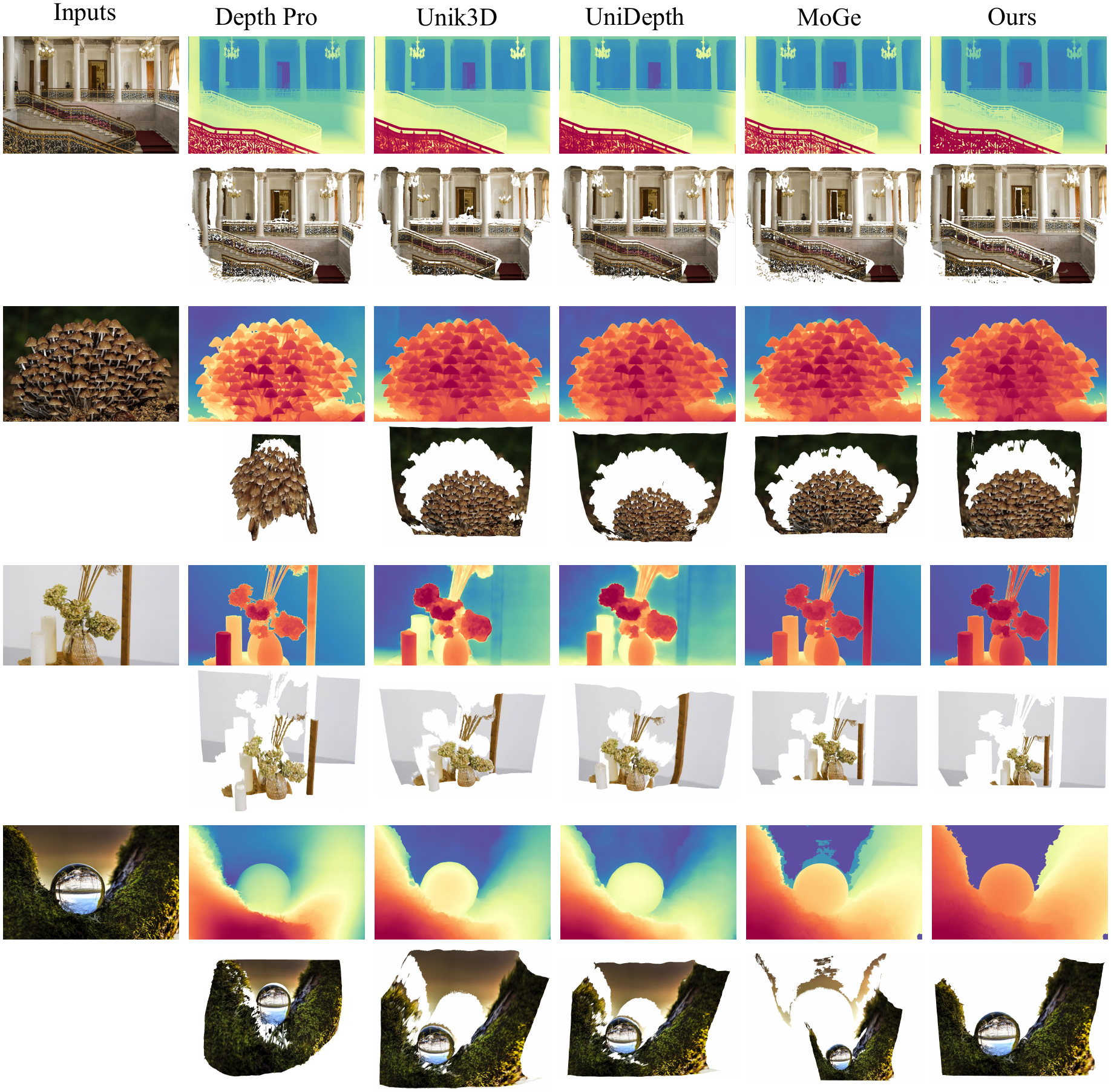}
    \caption{\textbf{Qualitative comparison on perspective scenes.} Compared with Depth Pro, UniK3D, UniDepth, and MoGe, DepthMaster demonstrates a strong capability in preserving fine-grained details and geometric structures, as evidenced by both the depth maps and 3D perspective views.}
    \label{fig:vis_pers_suppl}
\end{figure*}

\section{Limitations and Future Directions}\label{Appendix:limitations}

Following our discussions in the Section of Conclusion of the main paper, here we provide more analysis of DepthMaster's limitations, framing them as key directions for future research.

Our model shows limited generalization to highly abstract out-of-distribution (OOD) data. Although DepthMaster demonstrates strong generalization capabilities across various stylized scenes, it still struggles with highly abstract artistic styles, such as ink wash paintings. As illustrated in Figure~\ref{fig:limitation_abstract}, when confronted with such images, not only our DepthMaster but also all other state-of-the-art methods—including Depth Pro, UniK3D, UniDepth, and MoGe—fail to perceive the underlying 3D geometry, erroneously predicting them as flat planes. As a promising future direction, we plan to leverage image editing and style transfer models to synthesize a large-scale dataset featuring diverse abstract styles. Training on such constructed data will specifically enhance the model's depth perception capabilities in non-photorealistic and highly abstract domains.

\begin{figure*}[t!]
    \centering
    \includegraphics[width=\linewidth]{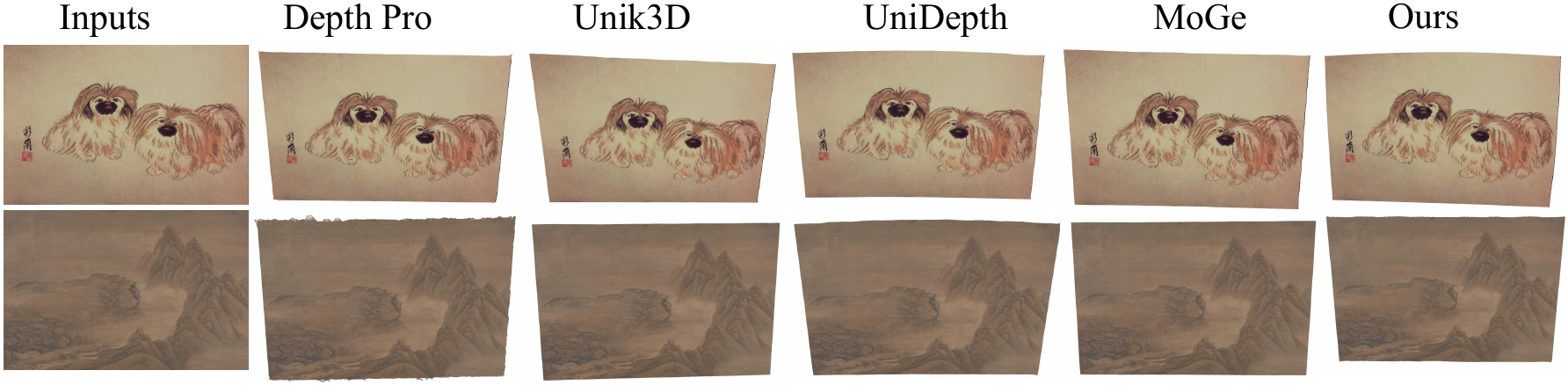}
    \caption{\textbf{Limitation on highly abstract artistic images.} For abstract styles such as ink wash paintings, our method, along with state-of-the-art methods (Depth Pro, UniK3D, UniDepth, and MoGe), fails to perceive the 3D geometry and incorrectly predicts the scenes as flat planes.}
    \label{fig:limitation_abstract}
\end{figure*}

\end{document}